\documentclass{article}
\usepackage{fancyhdr}

\usepackage{report}

\usepackage{soul}
\usepackage[utf8]{inputenc} 
\usepackage[T1]{fontenc}    
\usepackage{hyperref}       
\usepackage{url}            
\usepackage{booktabs}       
\usepackage{amsfonts}       
\usepackage{fancyhdr}       

\usepackage{nicefrac}       
\usepackage{microtype}      
\usepackage{graphicx}
\usepackage{pifont}
\usepackage{multirow}
\usepackage{CJKutf8}
\definecolor{darkmagenta}{rgb}{0.56, 0.0, 1.0}
\definecolor{softyellow}{rgb}{1.0, 0.92, 0.3} 
\definecolor{LightAquamarine}{rgb}{0.75, 1.0, 0.8} 
\definecolor{FireBrick}{RGB}{178,34,34}
\definecolor{MediumPurple}{RGB}{147,112,219}

\definecolor{uclablue}{rgb}{0.15, 0.45, 0.68}

\definecolor{headerblue}{RGB}{40, 80, 140}
\definecolor{headerbg}{RGB}{220, 235, 255}
\definecolor{closedtag}{RGB}{245, 240, 255}
\definecolor{opentag}{RGB}{240, 255, 240}
\definecolor{rowgray}{RGB}{248, 248, 252}
\definecolor{highlight}{RGB}{220, 190, 255}
\definecolor{best}{RGB}{0, 140, 100}
\definecolor{second}{RGB}{30, 100, 180}

\definecolor{genhead}{RGB}{52, 110, 183}
\definecolor{genbg}{RGB}{227, 238, 252}
\definecolor{edithead}{RGB}{46, 139, 87}
\definecolor{editbg}{RGB}{228, 248, 235}
\definecolor{repairhead}{RGB}{192, 72, 54}
\definecolor{repairbg}{RGB}{252, 232, 228}
\definecolor{rowalt}{RGB}{248, 249, 252}

\newcommand{\best}[1]{\textcolor{best}{\textbf{#1}}}
\newcommand{\second}[1]{\textcolor{second}{\underline{#1}}}

\definecolor{revblue}{RGB}{0,45,120}

\newif\ifshowrevisions
\showrevisionstrue

\hypersetup{
    breaklinks,
    colorlinks=true,
    citecolor={darkmagenta},
    linkcolor={uclablue},
    urlcolor={uclablue}
}
\usepackage{wrapfig}
\usepackage{float}
\usepackage{subcaption}
\usepackage{placeins}
\usepackage{lipsum} 
\usepackage{tcolorbox}
\usepackage{fvextra}
\usepackage{amsmath}
\usepackage{amssymb}
\usepackage{utfsym}
\usepackage{fontawesome}
\usepackage{xspace}
\usepackage{enumitem}
\usepackage{multirow}
\usepackage{algorithm}
\usepackage{algorithmic}
\tcbuselibrary{most,listings}
\usepackage{enumitem}
\usepackage{colortbl}
\usepackage{fancyhdr}

\usepackage{transparent}

\definecolor{promptbg}{RGB}{248,249,252}
\definecolor{promptframe}{RGB}{52,110,183}
\definecolor{successbg}{RGB}{240,250,242}
\definecolor{successframe}{RGB}{60,140,80}
\definecolor{failbg}{RGB}{252,242,242}
\definecolor{failframe}{RGB}{180,60,60}
\definecolor{instructionbg}{RGB}{246,246,247}
\definecolor{instructionframe}{RGB}{140,140,140}
\definecolor{diffbg}{RGB}{250,250,250}
\definecolor{diffframe}{RGB}{120,120,120}
\definecolor{diffadd}{RGB}{40,130,60}
\definecolor{diffrem}{RGB}{180,60,60}

\tcbset{
  promptboxbase/.style={%
    listing only,
    fonttitle=\small\bfseries,
    listing options={%
      basicstyle=\ttfamily\scriptsize,
      breaklines=true,
      breakautoindent=false,
      breakindent=0pt,
      postbreak=\mbox{\textcolor{gray}{$\hookrightarrow$}\space},
      escapeinside={(*@}{@*)},
      columns=fullflexible,
      keepspaces=true,
      aboveskip=0pt,
      belowskip=0pt,
      extendedchars=true,
      inputencoding=utf8,
      literate=%
        {—}{{---}}3
        {–}{{--}}2
        {‘}{{`}}1
        {’}{{'}}1
        {“}{{``}}2
        {”}{{''}}2
        {…}{{...}}3,
    },
    boxrule=0.6pt,
    arc=2pt,
    left=6pt, right=6pt, top=2pt, bottom=2pt,
    breakable,
  },
}

\newtcblisting{promptbox}[1]{%
  promptboxbase,
  colback=promptbg,
  colframe=promptframe,
  title={\small\bfseries #1},
}

\newtcblisting{successbox}[1]{%
  promptboxbase,
  colback=successbg,
  colframe=successframe,
  title={\small\bfseries\faCheck\space #1},
}

\newtcblisting{failbox}[1]{%
  promptboxbase,
  colback=failbg,
  colframe=failframe,
  title={\small\bfseries\faTimes\space #1},
}

\newtcolorbox{diffbox}[1]{%
  colback=diffbg,
  colframe=diffframe,
  title={\small\bfseries\faExchange\space #1},
  fonttitle=\small\bfseries,
  boxrule=0.6pt,
  arc=2pt,
  left=6pt, right=6pt, top=2pt, bottom=2pt,
  breakable,
}

\newtcolorbox{instructionbox}{%
  colback=instructionbg,
  colframe=instructionframe,
  boxrule=0.6pt,
  arc=2pt,
  left=6pt, right=6pt, top=3pt, bottom=3pt,
  breakable,
}

\definecolor{njuPurple}{RGB}{220,205,230}     
\definecolor{njuPurpleLight}{RGB}{250,245,252}   

\newtcolorbox{abstractbox}{
    colback=njuPurpleLight,   
    colframe=njuPurple,       
    boxrule=1pt,              
    arc=4mm,                  
    left=8pt,                 
    right=8pt,                
    top=8pt,                  
    bottom=8pt,               
    opacityback=0.95
}

\title{MMG2Skill: Can Agents Distill In-the-Wild Guides into Self-Evolving Skills?}

\author{
  Xinyu Che$^{*}$ \quad Junqi Xiong$^{*}$ \quad Yunfei Ge$^{*}$ \quad Xinping Lei$^{*}$ \quad Shihao Li$^{*}$ \quad Hang Yan$^{*}$ \quad Han Li$^{1}$ \quad Yuanxing Zhang \quad Zhiqi Bai\quad Jinhua Hao \quad Ming Sun \\ Han Li$^{2}$ \quad
  Jiaheng Liu$^{\dagger}$
  \\
\vspace{4mm}
\large
\textbf{Nanjing University}, \textbf{Kuaishou Technology}
\\
\vspace{2mm}
\texttt{kosmoche@gmail.com},    \texttt{liujiaheng@nju.edu.cn} \\
}

\begin{document}

\maketitle
\let\oldthefootnote\thefootnote

\let\thefootnote\relax\footnotetext{*~Equal Contribution. ~~$^\dagger$~Corresponding Author. \\ \hspace*{1.8em}$^{1}$~Nanjing University. ~~$^{2}$~Kuaishou Technology.}
\let\thefootnote\oldthefootnote

\begin{abstractbox}
\begin{center}
\textbf{\Large Abstract}
\end{center}
Abundant procedural knowledge on the Web holds great potential for helping agents solve long-horizon tasks. However, such knowledge is often multimodal, heterogeneous, noisy, and implicitly assumes human executors, making it difficult to use directly as the skills required by agents. To bridge the gap between human-oriented guides and agent-executable skills, we formalize this problem as \emph{guide-to-skill learning}: converting in-the-wild guides into executable skills and continuously improving them from trajectories observable to the agent.
To evaluate the capability of existing agents on this task, we introduce \textbf{MMG2Skill-Bench}, the first benchmark designed for this problem. We further propose \textbf{MMG2Skill}\footnote{\url{https://github.com/NJU-LINK/MMG2Skill}.}, a closed-loop framework that compiles guides into editable skills, conditions a fixed vision-language model (VLM) agent on these skills during execution, and revises the skills from trajectory-level root-cause feedback without using benchmark scores.
Across GUI control, open-ended gameplay, and strategic card play with six VLM backbones, MMG2Skill consistently outperforms vanilla baseline agents in every model–domain setting, achieving macro-average gains of +12.8 to +25.3 percentage points across backbones. Ablation studies show that directly prompting agents with raw guides can degrade performance, while both structured skill construction and trajectory-driven revision are necessary for the observed improvements. On success-inferable tasks, analyzer-based early stopping further prevents late-stage performance regressions and saves 25\%–53\% of attempts when the success signal is properly calibrated.
\end{abstractbox}

\providecommand{\TODO}[1]{\textcolor{red}{[TODO: #1]}}

\section{Introduction}
\label{sec:intro}

Vision-language model (VLM) agents are increasingly expected to perform long-horizon procedural work in interactive environments, from operating software in desktop GUIs~\citep{xie2024osworld,tan2024cradle,agashe2025agent} to completing open-world game objectives~\citep{fan2022minedojo,yuan2023plan4mc} and playing rule-constrained card games~\citep{zha2021rlcard}. The main obstacle is not only perception or next-action reasoning, but also weak \emph{procedural grounding}, the ability to anchor reusable procedures in the agent's runtime state. Such procedures must tell an agent when a behavior applies, which intermediate states indicate progress, and how to recover when execution drifts, and they must persist as editable objects that the agent can consult, localize to the current state, and revise after failure. We use skills to refer to these executable and editable procedural objects. Hand-writing skills for every task is labor intensive, while skills discovered only through agent exploration are limited by the agent's own failures and can spend extra trials reconstructing procedural knowledge already available in human-authored guides.

\begin{figure*}[t]
\centering
\includegraphics[width=\linewidth]{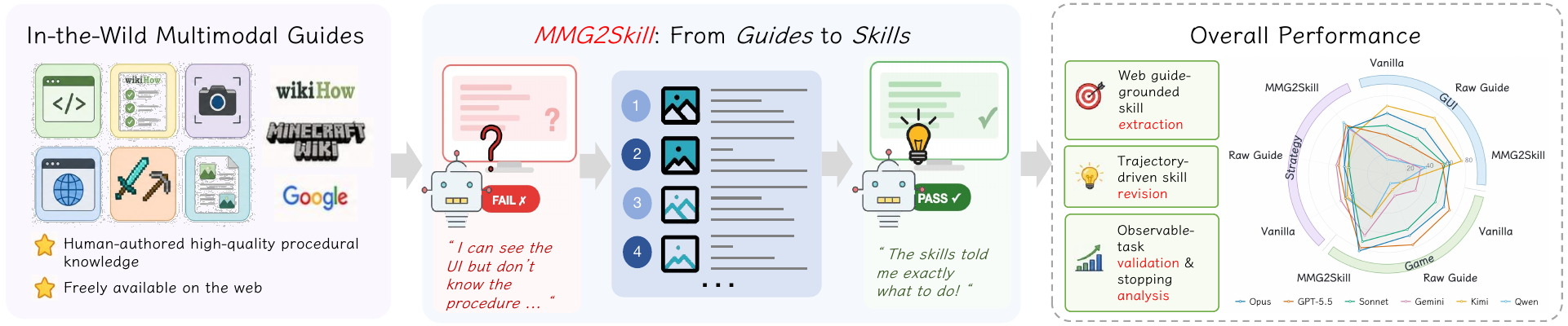}

\caption{\textbf{Motivation and high-level finding of MMG2Skill-Bench and MMG2Skill.} Public multimodal guides contain useful procedural knowledge for long-horizon agents, but raw guide prompting can hurt performance, motivating guide-to-skill learning.}
\label{fig:teaser}
\end{figure*}

Public Web guides provide this raw material at scale (Figure~\ref{fig:teaser}). Yet they are human artifacts rather than skills grounded in an agent's runtime state. They may contain sibling procedures, prerequisite assumptions, navigation text, and recovery advice that becomes wrong once the agent deviates from the nominal path, while the agent itself must still track which steps are already satisfied and align the rest with its current observation. The challenge is therefore not access to guides alone, but converting guides into compact skills that can be executed and locally revised against the agent's own trajectories. This raises the central question of this paper. \textbf{Can VLM agents distill in-the-wild guides into execution-grounded skills that improve through their own rollouts?}

Existing work leaves both the evaluation setting and the method incomplete. Context-learning benchmarks study whether models can use supplied information, but they usually use curated text and do not pair in-the-wild multimodal guides with environment-grounded execution~\citep{dou2026cl}. Skill benchmarks and self-improving agent methods study reusable skills, yet the skills are typically expert-provided, model-generated, or discovered through the agent's own experience~\citep{li2026skillsbench,wang2023voyager,zhao2024expel,shinn2023reflexion,ma2026skillclaw,zhang2026coevoskills}. As a result, we lack systematic evidence about whether public guide material can become agent-executable skills, and we lack a framework that can make this conversion without relying on benchmark scores during revision.

To address these gaps, we introduce \textbf{MMG2Skill-Bench}, the first benchmark for guide-to-skill learning from in-the-wild multimodal guides. The benchmark spans desktop GUI control, open-ended game play, and strategy tasks, enabling evaluation beyond a single environment or action interface. We further introduce \textbf{MMG2Skill}, a closed-loop framework named for converting \textbf{M}ulti\textbf{M}odal \textbf{G}uides to Skill. It compiles guides into editable skills, conditions a fixed VLM agent on the current skill set during rollout, and revises skills from agent-visible trajectories without using benchmark scores.

We evaluate MMG2Skill on MMG2Skill-Bench with six VLM backbones. MMG2Skill consistently outperforms vanilla agents in every model--domain setting, with macro-average gains of $+12.8$ to $+25.3$ percentage points across backbones (Figure~\ref{fig:teaser}).

We make two contributions. \textbf{(1)} MMG2Skill-Bench, the first benchmark pairing success-inferable interactive tasks with in-the-wild multimodal guides across GUI, Game, and Strategy domains. \textbf{(2)} MMG2Skill, a closed-loop framework that distills guides into editable \texttt{SKILL.md} procedures and revises them from trajectory-level root-cause feedback without benchmark scores.

Using the benchmark and framework, we identify three findings. \textbf{First}, public guides hold procedural knowledge that vanilla agents cannot reliably recover. \textbf{Second}, access alone is insufficient. Raw guide prompting can hurt, skill construction provides a safer procedural prior, and trajectory-driven revision repairs guide--runtime grounding gaps. \textbf{Third}, revision gains are non-monotonic, making calibrated stopping necessary for deployment. Analyzer-based early stopping mitigates late-stage regressions and saves 25--53\% of attempts on success-inferable tasks.

\providecommand{\TODO}[1]{\textcolor{red}{[TODO: #1]}}

\section{MMG2Skill-Bench}
\label{sec:benchmark}

MMG2Skill-Bench evaluates whether VLM agents can turn in-the-wild multimodal guides into procedural knowledge that remains useful during environment interaction, pairing each task instruction with task-relevant public guide material.

\subsection{Environments and Tasks}
\label{sec:benchmark:tasks}

MMG2Skill-Bench spans three interaction regimes. \textbf{MMG2Skill-GUI} uses OSWorld~\citep{xie2024osworld}, where agents operate desktop applications from screen observations and GUI actions. \textbf{MMG2Skill-Game} uses OpenHA Minecraft tasks~\citep{wang2025openha} in MineStudio~\citep{cai2024minestudio}, where agents complete open-ended objectives through exploration, crafting, and resource collection. \textbf{MMG2Skill-Strategy} uses Doudizhu and Mahjong tasks from RLCard~\citep{zha2021rlcard}, where agents make turn-based decisions from public observations and legal actions.

\begin{wrapfigure}{r}{0.48\textwidth}
\vspace{-1\baselineskip}
\centering
\includegraphics[width=\linewidth]{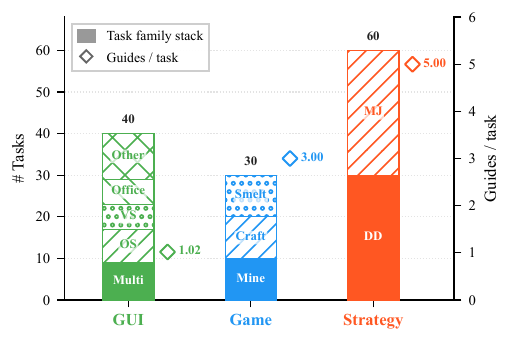}
\caption{\textbf{MMG2Skill-Bench composition.} Task-family counts and guide coverage by domain.}
\label{fig:benchmark_domain_guide_stack}
\vspace{-2\baselineskip}
\end{wrapfigure}

In total, MMG2Skill-Bench contains 130 success-inferable tasks (Figure~\ref{fig:benchmark_domain_guide_stack}). Appendix~\ref{app:task_selection} gives full task IDs and selection details.

\noindent\textbf{Scope of success-inferable evaluation.}
The main evaluation includes only tasks whose outcomes can be inferred from the agent-visible trajectory or a public final state. This keeps analyzer-based revision and stopping within the same information boundary as deployment. A separate No-Limit Hold'em set is used only as a private-information diagnostic because payoff depends on opponent private cards that may never appear in the agent-visible trajectory. We therefore exclude it from the main success-inferable Strategy evaluation and report it in Appendix~\ref{app:holdem_boundary}. Section~\ref{app:signal_calibration} quantifies the residual error in trajectory-only outcome inference.

\subsection{Guide Corpus}
\label{sec:benchmark:guides}

For each domain, we assemble public human-authored guide material and expose each task to a task-relevant subset. MMG2Skill-GUI draws on product documentation and how-to articles for desktop applications. MMG2Skill-Game draws on wiki pages and walkthrough-style guides that describe prerequisites, intermediate objectives, and resource procedures. MMG2Skill-Strategy uses public rule descriptions and beginner strategy material. The main representation pairs each page's HTML content with its image resources, preserving step order, optional visual examples, and interface or game-state cues.

\subsection{Evaluation Protocol and Controls}
\label{sec:benchmark:leakage}
\label{sec:benchmark:protocol}
Every agent, including the vanilla baseline, receives the same domain-specific system prompt before any guide is added. The prompt defines the executable interface, action grammar, observation format, control signals, and legal-action constraints where applicable. Guides supply the additional procedural knowledge to be converted into skills.

Although guides are task-relevant, they do not contain benchmark solution traces, gold action sequences, hidden evaluation labels, or environment-specific trajectories. We also exclude tasks whose answers can be copied directly from a public reference, so guides test procedural grounding rather than retrieval. Appendix~\ref{app:tutorials} details the per-domain guide sources and the corresponding leakage controls.

Each task is scored in $[0,1]$ using its domain-native evaluator. Benchmark-specific termination rules, step caps, opponents, and hyperparameters are deferred to Appendix~\ref{app:hparams}.

\providecommand{\TODO}[1]{\textcolor{red}{[TODO: #1]}}

\section{MMG2Skill Framework}
\label{sec:method}

\begin{figure*}[t]
    \centering
    \includegraphics[width=\linewidth]{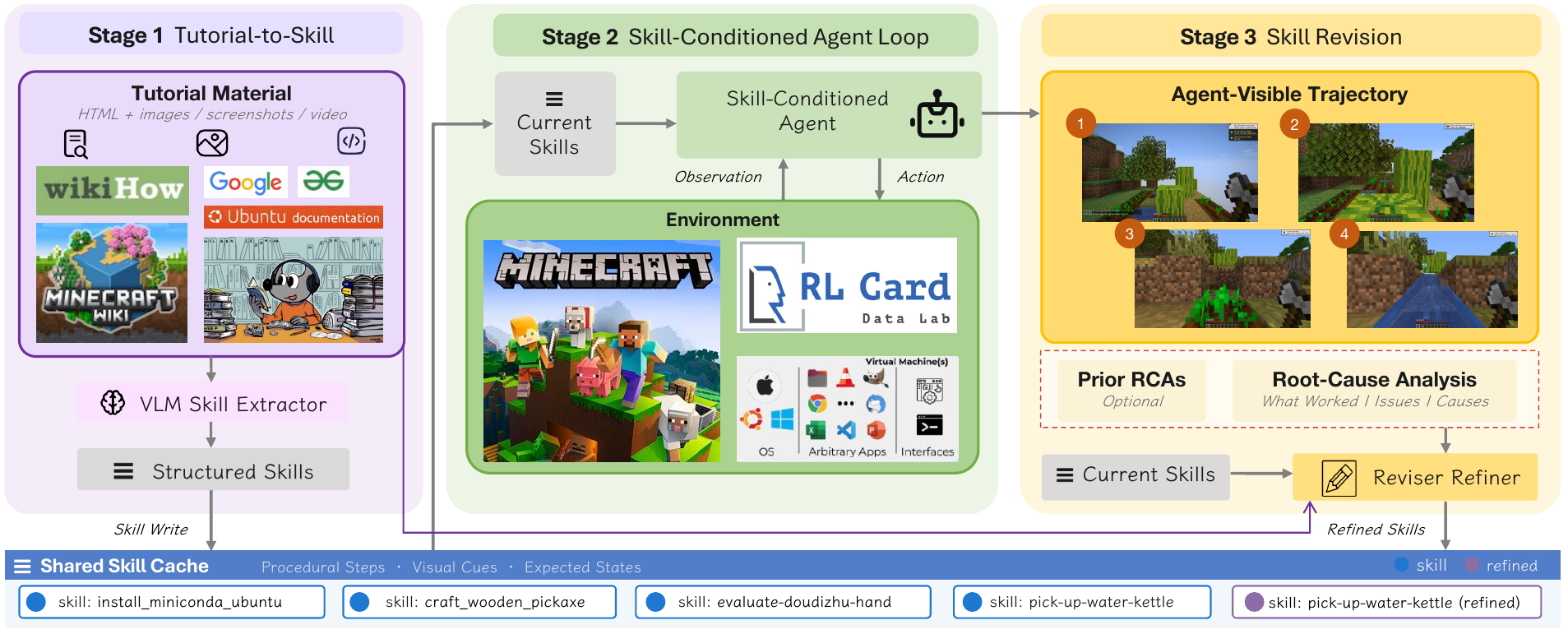}
    \caption{\textbf{MMG2Skill framework.} The figure shows the closed-loop pipeline that constructs guide-derived skills, uses them during rollout, and revises the skill cache from trajectory diagnoses across attempts.}
    \label{fig:method}
\end{figure*}

MMG2Skill compiles multimodal guides into editable skills and revises them from trajectory diagnoses across attempts (Figure~\ref{fig:method}).

\subsection{Problem Formulation}
\label{sec:method:formulation}

Let $I$ denote a task instruction and $G$ a multimodal guide. At step $t$, the agent observes $o_t$ and emits an action $a_t$. Because the VLM context is finite, the agent conditions on a bounded history $h_t$ that contains the current observation and the most recent $W{-}1$ preceding observation--action turns. A rollout is $\tau=(o_1,a_1,\ldots,o_L,a_L)$. The benchmark score $s(\tau)$ is used only for offline evaluation.

MMG2Skill keeps the VLM policy $\pi_\theta$ fixed across attempts, so $\theta$ is never updated. The optimization variable is an editable skill set $\mathcal{S} \in \Omega(G)$, where $\Omega(G)$ denotes skill sets induced from the guide. At the level of evaluation, the desired skill set is the one that maximizes expected score under the fixed policy:

\begin{equation}
    \mathcal{S}^{\star}
    \in
    \arg\max_{\mathcal{S} \in \Omega(G)}
    \mathbb{E}_{\tau \sim p_\theta(\tau \mid I, \mathcal{S})}
    \left[s(\tau)\right].
    \label{eq:skill_objective}
\end{equation}

Equation~\ref{eq:skill_objective} defines the evaluation target rather than the feedback available to the update operators. During construction, execution, analysis, and refinement, MMG2Skill never observes $s(\tau_k)$. It instead updates $\mathcal{S}_k$ using an analyzer diagnosis $\rho_k$, which summarizes agent-visible evidence about where the guide-derived skill set fails to match the rollout. Skills are initialized from $G$ rather than discovered from scratch. The construction and revision operators are instantiated in \S\ref{sec:method:framework}.

\subsection{Framework Stages}
\label{sec:method:framework}

\noindent\textbf{Multimodal skill construction.}
Human guides mix step order, screenshots, interface cues, and implicit assumptions that are difficult to use as raw context. MMG2Skill normalizes a multimodal guide into an editable \texttt{SKILL.md} representation.

\begin{equation}
\begin{aligned}
    \mathcal{S}_1 &= \textsc{ConstructSkills}(G, I) = \{z_i\}_{i=1}^{m}, \\
    z_i &= (u_i, c_i, v_i, q_i),
\end{aligned}
\label{eq:skill_construction}
\end{equation}

where $u_i$ is the reusable procedure, $c_i$ gives applicability conditions, $v_i$ records expected-state cues, and $q_i$ stores recovery knowledge. The tuple is a conceptual organization rather than a rigid schema. In implementation, skills express these fields through procedural text, state descriptions, and referenced guide images.

\noindent\textbf{Skill-conditioned execution.}
At the beginning of attempt $k$, the current skill set $\mathcal{S}_k$ is injected into the agent context and remains alongside the running history $h_t$ throughout the rollout. Every action is sampled under joint conditioning on the task, recent history, and current skills.

\begin{equation}
    a_t \sim \pi_\theta(a_t \mid h_t, I, \mathcal{S}_k).
    \label{eq:skill_policy}
\end{equation}

The skill set is added as procedural context while leaving each domain's action interface unchanged, so the same conditioning interface can be used across domains.

\noindent\textbf{Analyzer: converting rollouts into diagnoses.}
After trajectory $\tau_k$, the analyzer reads only the task instruction and agent-visible trajectory. It has no access to the benchmark score or hidden environment state. It then produces a diagnosis

\begin{equation}
    \rho_k = \textsc{Analyze}(I, \tau_k) = (e_k, r_k),
    \label{eq:framework_analysis_step}
\end{equation}

where $e_k$ is trajectory-grounded evidence about what worked, what failed, and where the behavior diverged from the task goal. The term $r_k$ is the analyzer's self-judged outcome assessment, with \texttt{likely\_success} indicating that the visible trajectory appears to satisfy the task. The refiner combines this diagnosis with the guide and current skills to produce localized skill edits, while $r_k$ provides the candidate stopping signal for deployment.

\noindent\textbf{Refiner: converting diagnoses into skill edits.}
Given the original guide, current skills, and accumulated diagnoses, the refiner updates the skill interface. At attempt $k$, MMG2Skill applies

\begin{equation}
    \mathcal{S}_{k+1} = \textsc{Refine}(G, I, \mathcal{S}_k, \rho_{1:k}).
    \label{eq:framework_refinement_step}
\end{equation}

The diagnosis chain $\rho_{1:k}$ accumulates without truncation, allowing the refiner to preserve earlier fixes when later attempts reveal new failures. The original guide $G$ remains part of the refinement input so that omitted or misunderstood procedural details can be recovered. The refiner edits only the skill representation, such as adding missing checks, sharpening state cues, reinforcing successful behavior, or removing misleading recovery advice.

\begin{algorithm}[t]
\caption{MMG2Skill closed-loop revision.}
\label{alg:mmg2skill_closed_loop}
\begin{algorithmic}[1]
\STATE $\mathcal{S}_1 \leftarrow \textsc{ConstructSkills}(G, I)$
\FOR{$k = 1$ \TO $N$}
    \STATE $\tau_k \leftarrow \textsc{Execute}(I, \mathcal{S}_k)$
    \STATE $\rho_k = (e_k, r_k) \leftarrow \textsc{Analyze}(I, \tau_k)$
    \IF{$r_k = \texttt{likely\_success}$}
        \STATE \textbf{return} $\mathcal{S}_k, \tau_k$
    \ENDIF
    \STATE $\mathcal{S}_{k+1} \leftarrow \textsc{Refine}(G, I, \mathcal{S}_k, \rho_{1:k})$ if $k < N$
\ENDFOR
\STATE \textbf{return} $\mathcal{S}_N, \tau_N$
\end{algorithmic}
\end{algorithm}

Algorithm~\ref{alg:mmg2skill_closed_loop} gives the deployment skeleton. The loop revises the skill set after each non-successful attempt and stops when the analyzer judges the visible trajectory as \texttt{likely\_success}. For offline policy comparisons, we additionally continue traces beyond would-be stopping points and apply the stopping rule retrospectively, which lets early-stop and full-run views be compared on the same attempt set. Appendix~\ref{app:algorithms} gives the complete procedure, including chunk-based analysis and history-aware refinement.

\providecommand{\TODO}[1]{\textcolor{red}{[TODO: #1]}}
\providecommand{\oldrev}[1]{\ifshowrevisions{\color{red}#1}\fi}

\section{Experiments}
\label{sec:exp}

We evaluate MMG2Skill on MMG2Skill-Bench (\S\ref{sec:benchmark}) using a single-backbone closed loop and organize the analysis around four research questions.

\begin{itemize}[leftmargin=*,topsep=2pt,itemsep=1pt]
    \item \textbf{RQ1: Overall performance} (\S\ref{sec:exp:main}). Do guide-derived skills improve a fixed VLM agent across the GUI, Game, and Strategy domains?
    \item \textbf{RQ2: Mechanism ablation} (\S\ref{sec:exp:ablation}). Where do the gains come from --- structured extraction, trajectory-driven revision, or their combination?
    \item \textbf{RQ3: Revision dynamics across attempts} (\S\ref{sec:exp:revision}). How does performance evolve across revision attempts?
    \item \textbf{RQ4: Early-stop deployment} (\S\ref{sec:exp:early_stop}). Is analyzer-based early stopping a reliable deployment choice compared with full-run deployment?
\end{itemize}

\subsection{Setup}
\label{sec:exp:setup}

We evaluate on the success-inferable tasks of MMG2Skill-Bench using the guide corpus and domain-native scoring protocol described in \S\ref{sec:benchmark}. Scores are reported as per-task scores multiplied by 100; differences use percentage points (pp). The private-information boundary case is reserved for an appendix diagnostic.

The main tables report the analyzer-selected early-stop view for MMG2Skill, which returns the first attempt within budget $N$ whose analyzer outcome is \texttt{likely\_success}, falling back to attempt $N$ if none triggers.

We compare MMG2Skill with two non-revising baselines. \texttt{vanilla} runs the same agent without external skills. \emph{Raw Guide} gives the agent the raw guide material directly but skips skill construction and disables the refiner. For MMG2Skill, each listed VLM backbone runs skill construction, execution, analysis, and refinement with the same model, without a separate stronger judge.

We use an attempt budget of $N{=}5$ for all main results, and additionally verify a larger budget by running Qwen3.6-Plus to $N{=}7$ (Appendix~\ref{app:extended_horizon}). Model variants and hyperparameters are in Appendices~\ref{app:model_card} and~\ref{app:hparams}.

\subsection{RQ1: Overall Performance}
\label{sec:exp:main}

\begin{table*}[t]
\centering
\renewcommand{\arraystretch}{1.15}
\resizebox{\textwidth}{!}{%
\begin{tabular}{@{}l*{12}{c}@{}}
\toprule
\multirow{2}{*}{\textbf{Model}} & \multicolumn{3}{c}{\textbf{GUI} (\%)} & \multicolumn{3}{c}{\textbf{Game} (\%)} & \multicolumn{3}{c}{\textbf{Strategy} (\%)} & \multicolumn{3}{c}{\textbf{Avg} (\%)} \\
\cmidrule(lr){2-4} \cmidrule(lr){5-7} \cmidrule(lr){8-10} \cmidrule(lr){11-13}
 & Van. & RG & MMG2Skill & Van. & RG & MMG2Skill & Van. & RG & MMG2Skill & Van. & RG & MMG2Skill \\
\midrule
Claude-Opus-4.6        & \second{62.22} & \second{59.75} & \second{64.75} & \second{66.67} & \second{66.67} & \best{83.33}   & 41.67          & 44.17          & 60.83          & \best{56.85}   & \best{56.86}   & \best{69.64}   \\
GPT-5.5                & 39.74          & 37.24          & 57.24          & \best{73.33}   & \best{76.67}   & \second{80.00} & \second{49.17} & \second{50.00} & \second{65.00} & \second{54.08} & \second{54.64} & \second{67.41} \\
Claude-Sonnet-4.6      & 49.76          & 47.50          & 59.76          & 56.67          & 60.00          & 73.33          & 40.00          & 39.17          & 61.67          & 48.81          & 48.89          & 64.92          \\
Kimi-K2.6              & \best{69.75}   & \best{74.75}   & \best{77.25}   & 16.67          & 16.67          & 46.67          & 46.67          & 40.83          & 52.50          & 44.36          & 44.08          & 58.81          \\
Gemini-3.1-Pro-Preview & 19.97          & 19.99          & 35.00          & 33.33          & 23.33          & 66.67          & \best{54.17}   & \best{52.50}   & 60.83          & 35.82          & 31.94          & 54.17          \\
Qwen3.6-Plus           & 15.00          & 17.50          & 39.99          & 16.67          & 10.00          & 46.67          & 48.33          & 44.17          & \best{69.17}   & 26.67          & 23.89          & 51.94          \\
\bottomrule
\end{tabular}%
}
\caption{\textbf{Main results across the three MMG2Skill-Bench domains.} \best{Best} and \second{second-best} marked per column. Van.\ = Vanilla, RG = Raw Guide.}
\label{tab:main_results_long}
\end{table*}

\noindent\textbf{MMG2Skill improves every cell.}
Under the main HTML+image guide representation, every one of the $18$ model--domain cells in Table~\ref{tab:main_results_long} yields a positive gain over vanilla, with the largest single-cell improvement reaching $+33.33$\,pp (Gemini on the Game domain). Appendix~\ref{app:full_results} shows the same pattern under the rendered-screenshot guide variant, indicating that the effect does not depend on a single webpage representation.

\noindent\textbf{Benefits span backbone strength.}
Absolute gains are largest from low vanilla starting points (Qwen on the GUI domain improves by $+25$\,pp), and strong backbones still benefit (GPT-5.5 on the Game domain, $+6.67$\,pp). This pattern suggests that guide-derived skills provide procedural knowledge that is not redundant with what stronger backbones already internalize. A representative Game-domain instance is bamboo-to-sticks crafting (Appendix~\ref{app:case_bamboo}). When the task supplies bamboo as the raw material, vanilla agents still attempt the wooden-plank recipe and exhaust their step budget searching for wood, whereas MMG2Skill agents follow the guide-derived bamboo path. This shows that public procedural knowledge is not necessarily available in the agent's task prior.

\noindent\textbf{Gains do not come from longer rollouts.}
In GUI and Game, MMG2Skill improves scores while reducing displayed-attempt steps on average, suggesting fewer exploratory detours rather than longer context-driven rollouts (Appendix~\ref{app:step_efficiency}).

\subsection{RQ2: Mechanism Ablation}
\label{sec:exp:ablation}

Table~\ref{tab:skill_extraction_ablation} compares raw guide injection, structured extraction without revision, and the full closed loop.

\noindent\textbf{Raw guide injection is unreliable.}
Compared with vanilla, injecting the raw guide is essentially flat on the GUI domain but \emph{decreases} performance on the Game and Strategy domains, with the largest loss at $-1.67$\,pp on Game. The result points to a guide--environment grounding mismatch rather than missing procedural content. Human-facing guides often include sibling tasks, implicit starting-state assumptions, and steps that may no longer apply after runtime deviations. Because raw injection has no editable interface for resolving these mismatches, misleading fragments can dominate the useful procedural information.

\begin{wraptable}{r}{0.48\textwidth}
\centering
\renewcommand{\arraystretch}{1.15}
\resizebox{\linewidth}{!}{%
\begin{tabular}{@{}lcccc@{}}
\toprule
\textbf{Domain} & \textbf{Vanilla} & \textbf{Raw Guide} & \textbf{w/o revision} & \textbf{MMG2Skill} \\
\midrule
GUI      & 42.74 & 42.79 & 51.08 & \best{55.67} \\
Game     & 43.89 & 42.22 & 43.89 & \best{66.11} \\
Strategy & 46.67 & 45.14 & 47.92 & \best{61.67} \\
\bottomrule
\end{tabular}%
}
\caption{\textbf{Skill extraction ablation.} Domain-level paired averages. \emph{w/o revision} is the attempt-1 score of MMG2Skill with skill construction only.}
\label{tab:skill_extraction_ablation}
\end{wraptable}

\noindent\textbf{Gains split along a guide--runtime grounding gradient.}
The \emph{w/o revision} variant matches or exceeds vanilla on all three domains, confirming that \textsc{ConstructSkills} turns guide content into a safer procedural prior. On GUI, one-shot extraction alone captures most of the total gain because human-written instructions translate cleanly into command-like or CLI-style operations, although screen actions still require observation-specific grounding. On Game and Strategy, revision instead contributes over $90\%$ of the total gain, with the largest single-stage contribution at $+22.22$\,pp on Game. Game guides specify recipes and workflows but leave inventory state, crafting interfaces, and spatial interaction to be resolved at runtime, while Strategy guides mostly provide high-level heuristics that must be rewritten as hand-specific decision rules. \texttt{SKILL.md} therefore matters as an editable interface, not as compressed guide context.

\noindent\textbf{Skill edits beat in-context feedback.}
A natural Reflexion-style alternative keeps the raw guide and feeds accumulated root causes back in context. Under a matched budget (same $N{=}5$, early-stop policy, and analyzer), we only vary whether root causes are persisted as \texttt{SKILL.md} edits or carried as in-context history (Appendix~\ref{app:raw_guide_root_cause}). It still trails MMG2Skill at $N{=}5$ for Kimi-K2.6 and Qwen3.6-Plus on GUI and Game, indicating that root-cause feedback contributes most when it is materialized as persistent edits to \texttt{SKILL.md} rather than accumulated as ephemeral context.

\subsection{RQ3: Revision Dynamics across Attempts}
\label{sec:exp:revision}

\begin{figure*}[t]
\centering
\includegraphics[width=\textwidth]{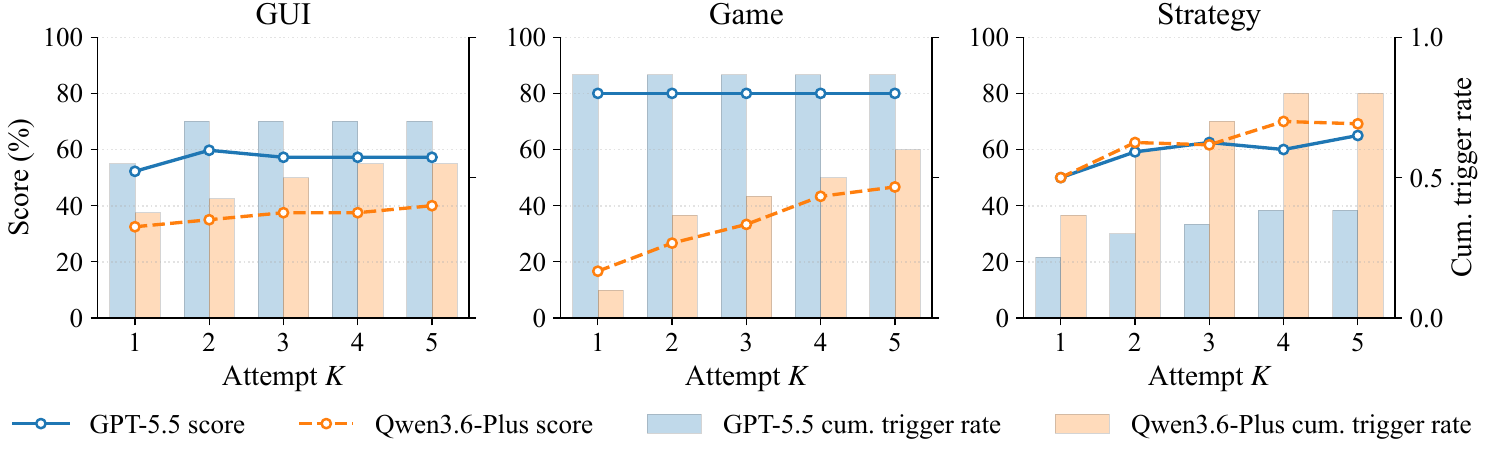}
\caption{\textbf{Revision dynamics across attempts $K$.} Lines show per-attempt mean score (\%, left axis) for GPT-5.5 and Qwen3.6-Plus on the three domains. Bars show cumulative \texttt{likely\_success} trigger rate (right axis).}
\label{fig:revision_dynamics}
\end{figure*}

\noindent\textbf{Revision gains are broad but uneven.}
Across the $N{=}5$ budget, the six-model averaged early-stop score lifts on every success-inferable domain, but the shape of the rise differs by both domain and backbone strength. Figure~\ref{fig:revision_dynamics} traces per-attempt dynamics for GPT-5.5 and Qwen3.6-Plus, the two models that bracket the per-model range, with model-averaged trajectories in Appendix~\ref{app:signal_calibration}. GPT-5.5 starts with stronger first-attempt performance on the GUI and Game domains, with its remaining across-attempt gains concentrated mainly on Strategy. Qwen3.6-Plus starts much lower and accrues most of its gain through revision on Game and Strategy. The two profiles point at the same mechanism, where the agent has weaker priors, revision contributes more of the across-attempt gain.

\noindent\textbf{Revision repairs guide--runtime mismatches.}
Static extraction cannot observe these mismatches because the guide may remain correct in its original human-facing context. A representative GUI-domain GIMP crop case in Appendix~\ref{app:case_gimp_crop} shows this mechanism. The initial skill contains the nominal crop procedure, but the failed rollout reveals that the agent also needs an explicit output-file verification step before the task can be considered complete. Revision converts this trajectory evidence into a missing runtime contract.Similar mismatches recur in Game and Strategy, underpinning the revision gains in Figure~\ref{fig:revision_dynamics}.

\noindent\textbf{Taking the latest attempt is not always safer.}
These revision gains do not make the latest available attempt automatically preferable. The full-run view, which always takes the latest available attempt, can plateau or regress when the analyzer proposes corrections from ambiguous evidence, especially in Strategy where a single hand has higher variance than a GUI task. The same non-monotonic pattern recurs in per-model and extended-horizon traces (Appendices~\ref{app:per_model_revision} and~\ref{app:extended_horizon}). This motivates comparing deployment policies in RQ4.

\subsection{RQ4: Early-Stop Deployment}
\label{sec:exp:early_stop}

\begin{figure*}[t]
\centering
\includegraphics[width=\linewidth]{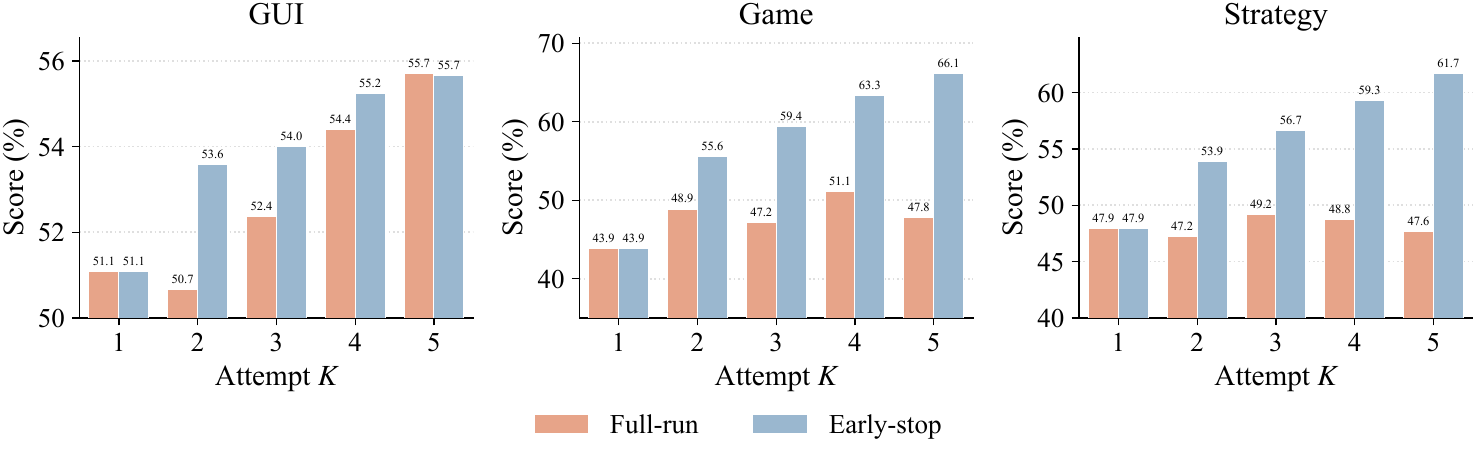}
\caption{\textbf{Policy gap between early-stop and full-run deployment.} Per-domain mean score (\%) at each attempt $K$, averaged across the model set. Bars contrast the full-run view (attempt~$K$) with the early-stop view (\S\ref{sec:exp:setup}).}
\label{fig:policy_gap}
\end{figure*}

For offline comparison, we replay the full-run and early-stop policies on the same recorded attempt traces, which continue beyond would-be stopping points as described after Algorithm~\ref{alg:mmg2skill_closed_loop}.

\begin{wraptable}{r}{0.48\textwidth}
\vspace{-1\baselineskip}
\centering
\renewcommand{\arraystretch}{1.15}
\resizebox{\linewidth}{!}{%
\begin{tabular}{@{}lccccc@{}}
\toprule
\textbf{Domain} & \textbf{P (\%)} & \textbf{R (\%)} & $\bar{k}$ & \textbf{Att. Save (\%)} & \textbf{Call Save (\%)} \\
\midrule
GUI      & 74.16 & 98.51 & 2.35 & 52.92 & 41.77 \\
Game     & 87.02 & 95.80 & 2.54 & 49.22 & 37.46 \\
Strategy & 84.56 & 70.79 & 3.73 & 25.44 & 23.60 \\
\bottomrule
\end{tabular}%
}
\caption{\textbf{Analyzer calibration and cost savings.} Precision and recall of the early-stop trigger. $\bar{k}$ is the mean stopping attempt; attempt and call savings are measured against the full-run budget (averaged per task).}
\label{tab:analyzer_signal_summary}
\vspace{-1\baselineskip}
\end{wraptable}

\noindent\textbf{Early stopping is the safer deployment view.}
On the core success-inferable domains (Figure~\ref{fig:policy_gap}), GUI nearly converges by $N{=}5$ ($55.7\%$ early-stop vs.\ $55.7\%$ full-run). On Game and Strategy, early-stop is substantially higher, with the largest gap on Game ($66.1\%$ vs.\ $47.8\%$ at $N{=}5$).

\noindent\textbf{Early stopping also saves attempts.}
The analyzer signal also reduces attempt count under online deployment, saving 25.44--52.92\% of attempts across domains relative to the full $N{=}5$ budget (Table~\ref{tab:analyzer_signal_summary}). The corresponding VLM API-call saving is 23.60--41.77\%, after accounting for fixed extraction, per-step agent calls, chunked analyzer calls, and refiner calls. These savings are measured against always running the full MMG2Skill budget, not against vanilla, because MMG2Skill adds analyzer and refiner calls.

\noindent\textbf{The signal is calibrated on success-inferable tasks.}
The analyzer's \texttt{likely\_success} assessment stays above $74\%$ precision on all three domains and above $95\%$ recall on GUI and Game, where outcomes are usually visible from the trajectory or public final state. Strategy remains precise ($84.6\%$) but has lower recall ($70.8\%$), indicating a conservative signal that misses some successful attempts. These results therefore support an early-stop policy for success-inferable tasks. In the Hold'em diagnostic, by contrast, outcome-determining cards may remain unobserved, so analyzer-based stopping should be disabled or replaced with external outcome feedback. The GUI essay-submission case in Appendix~\ref{app:case_essay_submission} illustrates the remaining false-positive pattern, where surface completion cues trigger the signal even though the oracle judge assigns score $0$. Appendix~\ref{app:signal_calibration} reports the full per-model breakdown.

\subsection{Residual Failures}
\label{sec:exp:failure_analysis}

\begin{figure*}[t]
\centering
\includegraphics[width=\linewidth]{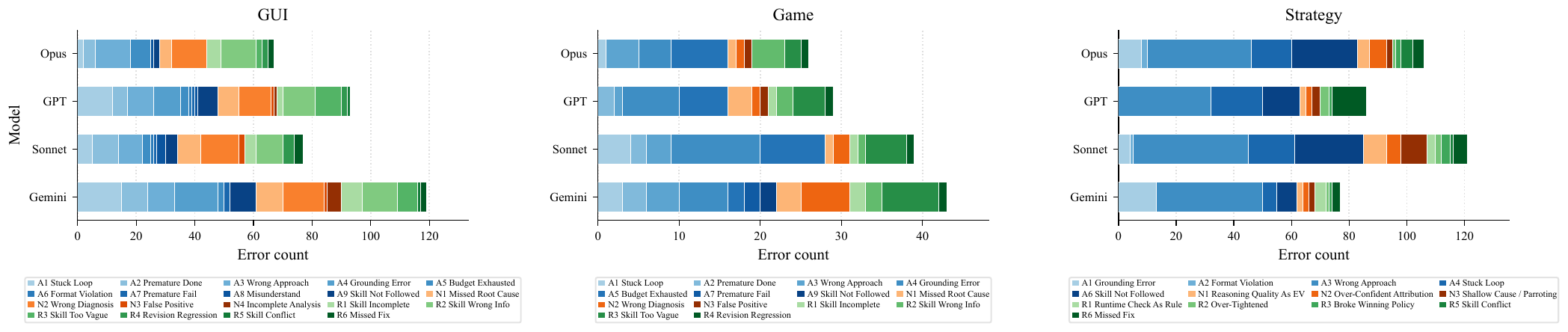}
\caption{\textbf{MMG2Skill failure attribution by pipeline component.} Each panel shows failed MMG2Skill cases in one domain, stacked by component and error-code category across the four backbones. Agent-loop skill-follow errors are placed last within the agent-loop taxonomy, and non-error analyzer or refiner tags are excluded.}
\label{fig:error_mmg2skill_components}
\end{figure*}

\noindent\textbf{Residual failures expose domain-specific bottlenecks.}
We annotate residual MMG2Skill failures with pipeline-component error tags, and read Figure~\ref{fig:error_mmg2skill_components} as a residual-failure diagnostic rather than a stage ranking. Each bar counts distinct error tags among cases that still fail under MMG2Skill. Remaining failures are distributed across the execution loop, analyzer, and refiner, but the meaning of the distribution changes by domain. GUI and Game failures still emphasize grounded interaction after a skill is available, whereas Strategy failures emphasize whether the model can convert a revised skill into the next policy choice. This split separates failures of following and executing a skill from failures of diagnosing and rewriting the skill, and it shows that the remaining failures are governed by each domain's execution regime rather than by a single pipeline component.

\noindent\textbf{GUI failures couple grounding, diagnosis, and skill information.}
On GUI, the residual profile is less concentrated in a single component than in the other domains. Opus has comparable counts for wrong approach, wrong diagnosis, and wrong skill information, which makes its remaining failures look like mismatches among the task intent, the trajectory explanation, and the revised instruction. GPT and Gemini show more stuck-loop and grounding-error tags, while Sonnet shows more premature completion alongside missed or wrong diagnoses. The within-domain commonality is that GUI failures are rarely only skill-follow failures. A model can receive a skill and still misread the screen state, choose a brittle action sequence, or carry an inaccurate diagnosis into the next revision.

\noindent\textbf{Game failures are execution-heavy.}
On Game, the residual profile is more execution-heavy. Opus, GPT, and Sonnet share a strong pattern of grounding errors and budget exhaustion, while Gemini shifts more of its residual mass toward wrong diagnosis and vague or wrong skill edits. Compared with GUI, analyzer errors occupy a smaller share of the bars, so the bottleneck is less often the act of explaining the failed trajectory. It is more often the act of completing a long grounded procedure before the attempt budget is exhausted. Within the domain, the common failure mode is therefore procedural fragility. Even after revision, the rollout can still fail when navigation, object interaction, or recovery behavior has to be executed precisely.

\noindent\textbf{Strategy failures hinge on decision-rule conversion.}
On Strategy, the domain shift is clearest. Failures are again dominated by the agent loop, but the dominant labels are wrong approaches, suboptimal strategy loops, grounding errors over the game state, and skill-not-followed cases rather than visuomotor execution errors. All four models retain wrong-approach errors as a major source of residual failure. Opus and Sonnet also show more skill-not-followed tags, GPT shows more stuck-loop tags, and Gemini shows more grounding-error tags relative to its skill-follow errors. Analyzer and refiner errors are smaller in count, but they are qualitatively specific to strategic revision. They include shallow causes, over-confident attribution, over-tightened rules, broken winning policies, and missed fixes. This pattern suggests that Strategy failures are not mainly caused by missing procedural details. They arise when a revised skill has to become a context-dependent decision rule rather than a fixed action recipe.

\section{Related Work}
\label{sec:related}

\noindent\textbf{Agent and context benchmarks.} AgentBench~\citep{liu2024agentbench}, Mind2Web~\citep{deng2023mind2web}, WebShop~\citep{yao2022webshop}, WebArena~\citep{zhou2024webarena}, VisualWebArena~\citep{koh2024visualwebarena}, OSWorld~\citep{xie2024osworld}, and MineDojo~\citep{fan2022minedojo} evaluate end-to-end agents in general, web, desktop, visual web, or embodied environments. CL-bench instead measures whether models can learn new knowledge, rules, and procedures from supplied textual contexts~\citep{dou2026cl,si2026context}. SkillsBench evaluates whether agents benefit from curated or self-generated skills under verifier-based tasks~\citep{li2026skillsbench}. MMG2Skill-Bench targets a different axis: each task is paired with public multimodal guide material, and methods are evaluated by whether they compile that material into persistent skills that improve interactive rollouts without benchmark-score feedback.

\noindent\textbf{Skills and procedural augmentation.} Recent agent systems augment LLMs with reusable behavior stored as language memories~\citep{zhao2024expel,shinn2023reflexion}, executable programs~\citep{wang2023voyager,liang2023code}, or skill libraries~\citep{ma2026skillclaw,zhang2026coevoskills}. Public tutorials and demonstrations have also been used as knowledge sources~\citep{fan2022minedojo}, reward signals, training data, or trajectory corpora~\citep{zhang2026tongui}. These systems show the value of procedural knowledge, but the reusable artifacts are usually learned from model rollouts, written by experts, or absorbed into training data. MMG2Skill initializes editable inference-time skills from public human-written guides and revises only the skill files from agent-visible trajectory diagnoses~\citep{madaan2023self}, without consuming benchmark scores.

\providecommand{\TODO}[1]{\textcolor{red}{[TODO: #1]}}
\providecommand{\oldrev}[1]{\ifshowrevisions{\color{red}#1}\fi}

\section{Conclusion}
\label{sec:conclusion}
We introduced \textbf{MMG2Skill-Bench}, a benchmark for evaluating whether VLM agents can turn in-the-wild multimodal guides into execution-grounded skills, and \textbf{MMG2Skill}, a closed-loop framework that compiles guides into editable skills and revises them from agent-visible trajectories. Across GUI control, open-world games, and strategic card play, MMG2Skill consistently improves performance on success-inferable tasks. The results show that guide access alone is insufficient. Raw guide prompting can hurt performance, whereas skill construction provides a safer procedural prior, and trajectory-driven revision repairs guide--runtime grounding gaps. Analyzer-based early stopping further improves deployability by avoiding late-stage regressions when the success signal is calibrated.

\section*{Limitations}
\label{sec:limitations}

The current study focuses on skill construction and revision after a task-relevant guide has been provided. This design isolates the question of whether multimodal procedural material can be turned into executable skills, but it does not study the upstream problem of guide discovery. In practice, an agent may need to search over multiple candidate sources, filter outdated or mismatched instructions, and decide when no guide is reliable enough to use. Our current system treats guides as static inputs selected before execution, so low-quality or misaligned guides can still introduce errors that revision only partly repairs. Extending MMG2Skill with retrieval and source filtering is a natural next step, but we leave this as a separate problem to keep the benchmark focused on execution-grounded skill learning.

Evaluating frontier VLM agents remains expensive. MMG2Skill requires multimodal API calls not only for task execution, but also for iterative skill revision and analyzer-based selection. As a result, continuously re-running the full benchmark on every newly released state-of-the-art model would be costly. We therefore report controlled comparisons under fixed model versions and evaluation budgets, and release prompts, metadata, and reproduction scripts to support future evaluations as model access and budgets permit.

Interactive GUI and simulator evaluation is slower than static text-only evaluation. Although our implementation parallelizes independent runs where possible, each rollout still involves sequential observation-action interaction with the environment. Evaluation throughput can also be constrained by commercial API rate limits, especially when multiple revision attempts are evaluated for each task. These constraints primarily affect evaluation turnaround time rather than the benchmark definition, and they motivate fixed protocols and explicit model-version records for reproducibility.

\section*{Ethical Considerations}
\label{sec:ethics}

MMG2Skill improves the ability of VLM agents to convert public procedural material into executable skills. This capability could be misused to automate unintended actions in real software environments. Our evaluation is restricted to sandboxed benchmark tasks and does not involve real user accounts, payment systems, private documents, or production services. Any credentials appearing in task instructions are synthetic benchmark artifacts. We do not recommend deploying revised skills in high-stakes or user-facing settings without external outcome verification, permission controls, and human oversight.

MMG2Skill-Bench builds on existing research benchmarks and software artifacts, including OSWorld, OpenHA and MineStudio, and RLCard. We use these artifacts for their intended research and evaluation purposes, cite their original sources, and keep their task boundaries separate from MMG2Skill's guide corpus and revision method. We do not redistribute benchmark assets beyond what their licenses or terms permit. OSWorld tasks are executed in its containerized desktop environment, while the game and card domains use benchmark simulators rather than real user accounts or production services.

MMG2Skill-Bench uses public web guides as procedural context. These sources include official documentation, community-maintained wikis, Q\&A pages, and third-party tutorials. Public accessibility does not imply permission to redistribute raw text or images. The release therefore accompanies guide artifacts with provenance metadata such as source URLs and source categories, and distinguishes cached contents from URL-only records. Raw guide contents are released only when redistribution is permitted by the corresponding source license or terms. For sources with unclear or restrictive redistribution terms, the release provides task-guide mappings, URLs, metadata, and collection scripts rather than cached raw contents. Released artifacts preserve attribution to original sources, and rights holders may request removal of cached guide contents from future releases.

We will release the code, benchmark metadata, prompts, reproduction scripts, and fixed RLCard opponent checkpoints as a public artifact, while excluding API keys, private logs, and third-party benchmark or guide assets whose redistribution is not permitted.

\bibliographystyle{unsrtnat}
\bibliography{references}

\appendix

\section{Responsible Research Details}
\label{app:responsible_research}

\noindent\textbf{AI assistance.}
We used AI writing assistance for language polishing, LaTeX checking, and code-debugging support. All scientific claims, experiments, analyses, and final text were reviewed by the authors.

\noindent\textbf{Human subjects.}
This study does not involve human-subject experiments, recruited participants, or private user data. Human effort is limited to benchmark construction, guide selection, and result inspection by the authors. The benchmark uses synthetic task instructions, public guide material, and domain-native scoring protocols.

\noindent\textbf{Supplementary materials.}
We provide anonymized code, benchmark metadata, prompts, checkpoint manifests, and reproduction scripts in the supplementary materials. Guide contents are included only when redistribution is permitted; otherwise, we provide task--guide mappings, source URLs, metadata, and collection scripts. The fixed RLCard opponent checkpoint weights are omitted from the submission package because of upload-size constraints, but will be released with the public artifact.

\section{Method and Implementation Details}

\subsection{Algorithm Details}
\label{app:algorithms}
\begin{algorithm}[htbp]
\caption{MMG2Skill revision loop with chunked analysis and history-aware refinement.}
\label{alg:mmg2skill_full}
\begin{algorithmic}[1]
\STATE \textbf{Input:} task instruction $I$, guide $G$, attempt budget $N$, chunk size $C$, mode $M \in \{\texttt{online}, \texttt{offline}\}$
\STATE \textbf{Output:} attempt records $\mathcal{R}$, final skill set $\mathcal{S}_K$, early-stop index $k_{\mathrm{stop}}$
\STATE $\mathcal{S}_1 \leftarrow \textsc{ConstructSkills}(G, I)$; $\mathcal{R} \leftarrow \emptyset$; $k_{\mathrm{stop}} \leftarrow \bot$
\FOR{$k = 1$ \TO $N$}
    \STATE $\tau_k \leftarrow \textsc{Execute}(I, \mathcal{S}_k)$
    \STATE \COMMENT{\textbf{Analyze}: chunked trajectory reading}
    \label{line:analyze_start}
    \STATE Parse $\tau_k$ into decision turns $\{t_1, \ldots, t_L\}$ and chunks $\{c_1, \ldots, c_{\lceil L/C \rceil}\}$
    \STATE $\sigma \leftarrow \text{``''}$
    \FOR{$i = 1$ \TO $\lceil L/C \rceil$}
        \IF{$i < \lceil L/C \rceil$}
            \STATE $\sigma \leftarrow \text{VLM}(I, \sigma, c_i; \textsc{ChunkSummary})$
        \ELSE
            \STATE $\rho_k \leftarrow \text{VLM}(I, \sigma, c_i; \textsc{RootCause})$
        \ENDIF
    \ENDFOR
    \STATE Parse $\rho_k = (e_k, r_k)$, where $e_k$ is trajectory evidence and $r_k$ is the analyzer outcome
    \label{line:analyze_end}
    \STATE $\mathcal{R} \leftarrow \mathcal{R} \cup \{(\tau_k, \rho_k)\}$
    \IF{$r_k = \texttt{likely\_success}$ and $k_{\mathrm{stop}} = \bot$}
        \STATE $k_{\mathrm{stop}} \leftarrow k$
        \IF{$M = \texttt{online}$}
            \STATE \textbf{return} $\mathcal{R}$, $\mathcal{S}_k$, $k_{\mathrm{stop}}$
        \ENDIF
    \ENDIF
    \IF{$k < N$}
        \STATE \COMMENT{\textbf{Refine}: history-aware skill rewriting}
        \label{line:refine_start}
        \STATE Render $\mathcal{S}_k$ as Markdown and render $G$ as text with images
        \STATE Assemble $\rho_{1:k}$ from the accumulated root-cause outputs
        \STATE Set edit intensity from $r_k$
        \STATE Reinforce successes, edit normally when uncertain, and allow larger edits for failures
        \STATE $\mathcal{S}_{k+1} \leftarrow \text{VLM}(I, G, \mathcal{S}_k, \rho_{1:k}; \textsc{RefineSkills})$
        \label{line:refine_end}
    \ENDIF
\ENDFOR
\STATE $K \leftarrow N$
\STATE \textbf{return} $\mathcal{R}$, $\mathcal{S}_K$, $k_{\mathrm{stop}}$
\end{algorithmic}
\end{algorithm}
Algorithm~\ref{alg:mmg2skill_full} expands the closed-loop procedure in \S\ref{sec:method:framework}. It makes two implementation details explicit. First, the analyzer reads long trajectories through fixed-size chunks and passes information across chunks only through a rolling summary. Second, the refiner rewrites the skill set using the full diagnosis history while never receiving benchmark scores. The algorithm supports both online deployment, where the first \texttt{likely\_success} diagnosis stops the loop, and offline diagnostics, where the loop continues to the full attempt budget so that early-stop and full-run views can be compared.

The chunked analysis step (lines~\ref{line:analyze_start}--\ref{line:analyze_end}) keeps each analyzer call within a fixed per-call context budget. Each intermediate call receives only the task instruction, the rolling summary, and the current trajectory chunk. The final call switches from summarization to root-cause analysis and emits the structured diagnosis used by both early stopping and refinement. The analyzer does not receive $\mathcal{S}_k$, hidden environment state, or the benchmark score $s(\tau_k)$.
The refinement step (lines~\ref{line:refine_start}--\ref{line:refine_end}) uses the original guide, the current skill set, and the accumulated diagnoses $\rho_{1:k}$. Keeping the full diagnosis history visible helps the refiner preserve earlier fixes when later attempts reveal new failures. The refiner outputs a normalized Markdown skill set rather than a local patch, so each skill remains represented once and validated behavior can be carried forward unchanged. Benchmark scores are added only to $\mathcal{R}$ during offline evaluation and are never included in analyzer or refiner inputs.

\subsection{Extensible Research Artifact}
\label{app:extensible_artifact}

The released code is designed to support follow-up work beyond the three environments studied in this paper. Alongside the benchmark data and prompts, we provide an adapter guide for adding new interactive domains to the same guide-to-skill evaluation pipeline. The goal is to make the core MMG2Skill loop reusable rather than tied to OSWorld, Minecraft, or RLCard. A new environment supplies the minimal domain boundary needed for interaction and evaluation, while skill construction, skill-conditioned execution, trajectory analysis, refinement, and early-stop reporting remain shared across domains.

This design supports two forms of reuse. First, researchers can evaluate alternative guide-to-skill methods under the same attempt budget, trajectory logging, and early-stop or full-run protocols. Second, they can instantiate new benchmark domains by pairing tasks with public guides and implementing the environment boundary, without rewriting the core skill-learning loop. We view this extensibility as important for studying whether guide-derived skills generalize across broader classes of interactive agents and procedural knowledge sources.

\subsection{Hyperparameters}
\label{app:hparams}
\begin{table*}[t]
\centering
\renewcommand{\arraystretch}{1.2}
\begin{tabular}{@{}lcccc@{}}
\toprule
\textbf{Parameter} & \textbf{GUI} & \textbf{Game}
 & \textbf{Strategy} & \textbf{Hold'em} \\
\midrule
\#tasks / \#hands                                & 40 & 30 & 60 & 150 \\
Max steps per rollout                            & 15 & 60 & 60 & 60 \\
History window $W$                               & 3  & 10 & 60 & 60 \\
Analyzer chunk size $C$                          & 15 & 15 & 30 & 30 \\
Attempt budget $N$                               & 5  & 5  & 5  & 5  \\
Skill max images                                 & 20 & 20 & 20 & 20 \\
\multirow{2}{*}{Termination}                     & max\_steps / & max\_steps / & is\_over() / & is\_over() / \\
                                                 & DONE / FAIL  & DONE / FAIL  & max\_steps   & max\_steps   \\
Opponent                                         & n/a & n/a & DMC / NFSP & DQN \\
\bottomrule
\end{tabular}
\caption{\textbf{Complete hyperparameter listing.} Settings referenced in \S\ref{sec:exp:setup} and Algorithm~\ref{alg:mmg2skill_full}. The Hold'em column is reported only as a private-information boundary diagnostic (Appendix~\ref{app:holdem_boundary}) and is not part of MMG2Skill-Bench's three main domains.}
\label{tab:appendix_hyperparameters}
\end{table*}

\noindent\textbf{Benchmark protocol.}
MMG2Skill-GUI is built on OSWorld, MMG2Skill-Game on Minecraft, and MMG2Skill-Strategy on the Doudizhu and Mahjong tasks from RLCard, with the No-Limit Hold'em diagnostic running on the same RLCard engine as MMG2Skill-Strategy. For MMG2Skill-GUI and MMG2Skill-Game, a rollout ends on a \texttt{DONE} token, an exhausted step budget, or a \texttt{FAIL} token. The \texttt{FAIL} token lets the agent abstain on OSWorld's infeasible-task subset, and is also exposed on MMG2Skill-Game. The step budgets are 15 for MMG2Skill-GUI and 60 for MMG2Skill-Game. For MMG2Skill-Strategy and the Hold'em diagnostic, each hand ends through the RLCard engine's natural termination (cards played out, a declared \emph{hu}, an exhausted wall, or a settled final betting round) under a shared cap of 60 VLM-call steps, and reaching the cap without natural termination is treated as a forfeit. No-Limit Hold'em uses the two-player heads-up setting with 100 chips per side. The opponents are trained RLCard agents. We use DMC for Doudizhu, NFSP for Mahjong, and DQN for No-Limit Hold'em.

\noindent\textbf{RLCard opponent training.}
All RLCard opponents are fixed during LLM evaluation and receive no trajectory feedback from MMG2Skill. We train them with RLCard's official implementations without modifying the environments. Doudizhu uses DMC with self-play over $10^8$ frames and saves seat-specific checkpoints for the landlord and two peasant seats. Mahjong uses NFSP for $5{\times}10^4$ episodes with seat 0 as the learner and RandomAgent in the other seats. No-Limit Hold'em uses DQN for $5{\times}10^4$ episodes in the two-player heads-up setting. DQN and NFSP use RLCard's default three-layer MLP with 128 hidden units per layer, Adam at $5{\times}10^{-5}$, batch size 32, and seed 42. DMC uses the default RLCard DMC trainer with 5 actors, RMSProp at $10^{-4}$, batch size 32, and exploration rate $\epsilon{=}0.01$. DQN and NFSP checkpoints are selected by training-time evaluation against RandomAgent every 100 episodes, while DMC uses the trainer's native checkpointing and is converted into seat-specific evaluation checkpoints. At LLM evaluation time, each opponent checkpoint is loaded on CPU and kept fixed.

\noindent\textbf{OSWorld model-specific interface handling.}
For MMG2Skill-GUI, we follow the official OSWorld execution protocol and its model-specific interface handling so that each backbone uses the interaction format expected by its provider or upstream adapter. Kimi receives normalized screen coordinates. Claude is evaluated at a forced $1280{\times}720$ desktop resolution. For PyAutoGUI typing actions, we follow the upstream OSWorld fix for newline and escaping behavior. Text is split across newline characters, \texttt{typewrite()} is called on each line, and \texttt{press('enter')} is issued between lines rather than typing the newline as literal text. The typed strings are also escaped before execution so quotes, backslashes, and other special characters are passed safely. These adjustments only affect action normalization and execution and are applied uniformly to all compared methods.

\noindent\textbf{MMG2Skill-Game attempt protocol.}
For each MMG2Skill-Game task ID we use a single canonical instruction across all attempts, selected so that the target item is reachable from the resources in the corresponding environment instance. The OpenHA instruction pool contains paraphrases that reference materials inconsistent with the assigned resource set, for example a planks-to-sticks paraphrase drawn for a task scoped to bamboo, so fixing the instruction also keeps each attempt solvable. We otherwise follow the OpenHA setup and resample the initial inventory between attempts, keeping the same item set but randomizing slot positions, so the revised skill must encode position-invariant procedures rather than memorize a fixed click sequence.

\noindent\textbf{MMG2Skill hyperparameters.}
We use an attempt budget of $N{=}5$ for every benchmark. All VLM calls use the default temperature and set the maximum output length to $32{,}768$ tokens. The analyzer trajectory chunk size is $C{=}15$ for MMG2Skill-GUI and MMG2Skill-Game and $C{=}30$ for MMG2Skill-Strategy (and the Hold'em diagnostic), reflecting their different rollout lengths. The Stage~2 history window is $W{=}3$ on MMG2Skill-GUI, where each step's observation is largely self-contained. We widen it to $W{=}10$ on MMG2Skill-Game so the agent can refer back to recent navigation and crafting actions whose effects only manifest several steps later. For MMG2Skill-Strategy and the Hold'em diagnostic we set $W{=}60$, equal to the rollout cap, so the agent always sees the full hand history when deciding the next action, since past plays and betting moves materially shape the optimal action. The Stage-1 \textsc{ConstructSkills} extractor accepts up to $20$ guide images per one-shot call (\emph{Skill max images} in Table~\ref{tab:appendix_hyperparameters}), so the initial skill construction has a generous visual budget without inflating downstream refiner context.

\subsection{Model Card}
\label{app:model_card}

Table~\ref{tab:model_card} lists the model variants referenced in the experiments. Within each backbone condition, the same model is used for skill construction, execution, analysis, and refinement.

\begin{table}[t]
\centering
\renewcommand{\arraystretch}{1.1}
\begin{tabular}{@{}l@{}}
\toprule
\textbf{Model} \\
\midrule
\texttt{claude-opus-4-6} \citep{anthropic2026opus46} \\
\texttt{gpt-5.5} \citep{openai2026gpt55} \\
\texttt{claude-sonnet-4-6} \citep{anthropic2026sonnet46} \\
\texttt{kimi-k2.6} \citep{kimik26} \\
\texttt{gemini-3.1-pro-preview} \citep{google2026gemini31pro} \\
\texttt{qwen3.6-plus} \citep{qwen36plus} \\
\bottomrule
\end{tabular}
\caption{\textbf{Model variants referenced in the experiments.} Experiments were completed in May 2026.}
\label{tab:model_card}
\end{table}

\section{Benchmark and Guide Corpus Details}
\subsection{Guide Materials}
\label{app:tutorials}

\noindent\textbf{Guide versus domain system prompt.}
The benchmark section states the prompt--guide boundary. The prompt supplies the executable interface, while the guide supplies procedural knowledge. Here we give the domain-specific split. MMG2Skill-GUI uses OSWorld prompts to provide the PyAutoGUI execution contract and control signals, while the guides provide application how-tos for tools such as LibreOffice, VS Code, and Chrome. MMG2Skill-Game uses Minecraft prompts to define the action vocabulary, coordinate conventions, GUI hazards, and success criteria. Guides describe how to compose those primitives into recipes, mining or smelting procedures, and inventory workflows. MMG2Skill-Strategy uses RLCard prompts to define player roles, turn structure, card-combination syntax, and legal-action constraints. Guides add strategic heuristics such as bomb timing, Peasant cooperation, betting thresholds, and chow/pong priorities.

\noindent\textbf{Guide selection and leakage controls.}
The exclusion list in \S\ref{sec:benchmark:leakage} is realized differently in each domain. MMG2Skill-Game uses target-item or related-entity Wiki pages that state recipes, uses, and possible sources without prescribing a route for a particular world state. MMG2Skill-GUI uses application documentation rather than benchmark answer scripts. MMG2Skill-Strategy uses general rules or beginner strategy rather than labels for particular hands.

\noindent\textbf{Pre-selection exclusion of guide-trivializable tasks.}
The exclusion in \S\ref{sec:benchmark:leakage} is specific to MMG2Skill-GUI in practice. Concrete dropped cases include listing conference host cities from a conference page and implementing an algorithm whose full solution is already in the guide. Both would test retrieval or copying rather than guide-to-action grounding.

Corpus statistics for the three MMG2Skill-Bench domains are reported in Table~\ref{tab:tutorial_stats}. The MMG2Skill-GUI set spans roughly two dozen distinct hosts. The main HTML representation stores the raw \texttt{page*.html} files plus referenced image assets, while the screenshot variant renders source pages as $1280{\times}2000$ rolling-window captures used in the appendix modality-robustness experiments.

\begin{table}[t]
\centering
\renewcommand{\arraystretch}{1.15}
\begin{tabular}{@{}lcccc@{}}
\toprule
\textbf{Domain} & \textbf{\#\,Guide} & \textbf{Guide/Task} & \textbf{Chars} & \textbf{Images} \\
\midrule
GUI      & 41 & 1.02 & 137.1k & 9.0  \\
Game     & 32 & 3.00 & 173.6k & 0.0  \\
Strategy & 10 & 5.00 & 452.9k & 18.0 \\
\bottomrule
\end{tabular}%
\caption{\textbf{Main guide corpus statistics for the three MMG2Skill-Bench domains.} \emph{Chars} and \emph{Images} are per-task means. See \S\ref{app:tutorials} for column definitions.}
\label{tab:tutorial_stats}
\end{table}

\begin{table*}[!t]
\centering
\renewcommand{\arraystretch}{1.2}
\scriptsize
\setlength{\tabcolsep}{4pt}
\begin{tabular}{@{}l l c >{\raggedright\arraybackslash}p{0.62\linewidth}@{}}
\toprule
\textbf{Domain} & \textbf{Task family} & \textbf{n} & \textbf{Task IDs or targets} \\
\midrule
\multirow[c]{10}{*}{\normalsize GUI}
 & \texttt{chrome}              & 3 & \texttt{2ae9ba84-3a0d-4d4c-8338-3a1478dc5fe3, 99146c54-4f37-4ab8-9327-5f3291665e1e, bb5e4c0d-f964-439c-97b6-bdb9747de3f4} \\
 & \texttt{gimp}                & 3 & \texttt{2a729ded-3296-423d-aec4-7dd55ed5fbb3, 554785e9-4523-4e7a-b8e1-8016f565f56a, b148e375-fe0b-4bec-90e7-38632b0d73c2} \\
 & \texttt{libreoffice\_calc}    & 2 & \texttt{4188d3a4-077d-46b7-9c86-23e1a036f6c1, 51719eea-10bc-4246-a428-ac7c433dd4b3} \\
 & \texttt{libreoffice\_impress} & 2 & \texttt{2cd43775-7085-45d8-89fa-9e35c0a915cf, 455d3c66-7dc6-4537-a39a-36d3e9119df7} \\
 & \texttt{libreoffice\_writer}  & 2 & \texttt{0b17a146-2934-46c7-8727-73ff6b6483e8, 0e47de2a-32e0-456c-a366-8c607ef7a9d2} \\
 & \texttt{multi\_apps}          & 9 & \texttt{2373b66a-092d-44cb-bfd7-82e86e7a3b4d, 48d05431-6cd5-4e76-82eb-12b60d823f7d, 510f64c8-9bcc-4be1-8d30-638705850618, 869de13e-bef9-4b91-ba51-f6708c40b096, 8df7e444-8e06-4f93-8a1a-c5c974269d82, 91190194-f406-4cd6-b3f9-c43fac942b22, e1fc0df3-c8b9-4ee7-864c-d0b590d3aa56, f7dfbef3-7697-431c-883a-db8583a4e4f9, f918266a-b3e0-4914-865d-4faa564f1aef} \\
 & \texttt{os}                  & 8 & \texttt{13584542-872b-42d8-b299-866967b5c3ef, 4783cc41-c03c-4e1b-89b4-50658f642bd5, 4d117223-a354-47fb-8b45-62ab1390a95f, 5c1075ca-bb34-46a3-a7a0-029bd7463e79, 94d95f96-9699-4208-98ba-3c3119edf9c2, a462a795-fdc7-4b23-b689-e8b6df786b78, b3d4a89c-53f2-4d6b-8b6a-541fb5d205fa, f9be0997-4b7c-45c5-b05c-4612b44a6118} \\
 & \texttt{thunderbird}         & 3 & \texttt{08c73485-7c6d-4681-999d-919f5c32dcfa, 3f28fe4f-5d9d-4994-a456-efd78cfae1a3, dfac9ee8-9bc4-4cdc-b465-4a4bfcd2f397} \\
 & \texttt{vlc}                 & 2 & \texttt{8ba5ae7a-5ae5-4eab-9fcc-5dd4fe3abf89, 8f080098-ddb1-424c-b438-4e96e5e4786e} \\
 & \texttt{vs\_code}             & 6 & \texttt{0512bb38-d531-4acf-9e7e-0add90816068, 70745df8-f2f5-42bd-8074-fbc10334fcc5, 9439a27b-18ae-42d8-9778-5f68f891805e, 982d12a5-beab-424f-8d38-d2a48429e511, c6bf789c-ba3a-4209-971d-b63abf0ab733, ea98c5d7-3cf9-4f9b-8ad3-366b58e0fcae} \\
\midrule
\multirow[c]{3}{*}{\normalsize Game}
 & \texttt{mine\_block} & 10 & \texttt{coal\_ore, cobblestone, dirt, gravel, iron\_ore, melon, oak\_log, pumpkin, sand, sugar\_cane} \\
 & \texttt{craft\_item} & 10 & \texttt{bookshelf, bread, chest, crafting\_table, light\_gray\_dye, loom, stick, sugar, torch, wheat} \\
 & \texttt{smelt\_item} & 10 & \texttt{brick, charcoal, cooked\_beef, dried\_kelp, gold\_ingot, green\_dye, iron\_nugget, smooth\_stone, sponge, terracotta} \\
\midrule
\normalsize Strategy
 & Doudizhu and Mahjong tasks & 60 & Fixed Strategy evaluation hands, with 30 Doudizhu hands and 30 Mahjong hands. \\
\bottomrule
\end{tabular}
\normalsize
\caption{\textbf{Task selection by MMG2Skill-Bench domain.} GUI lists the OSWorld application domains and UUIDs used in the evaluation. Game lists the MineStudio task families and target items. Strategy contains 60 RLCard hands drawn from Doudizhu and Mahjong tasks.}
\label{tab:task_selection}
\end{table*}

\noindent\textbf{MMG2Skill-Game HTML images.}
The MMG2Skill-Game HTML row reports zero images by design. A wiki page on \texttt{minecraft.wiki} ships large numbers of redundant icon assets. The same item appears at multiple sprite resolutions ($16{\times}16$, $32{\times}32$, and $64{\times}64$ icon files), and adjacent navigation/infobox panels embed icons for many tangentially related items that the agent does not need. In a rendered view and therefore in the screenshot variant, these icons collapse into a small number of in-flow images, but the raw HTML asset list is dominated by them. Because the typical sprite icon is too small to be meaningfully read by a VLM, we filter \emph{all} \texttt{<img>} references out of the MMG2Skill-Game HTML guides and feed the extractor text-only HTML. The screenshot pipeline is unaffected and retains the natural in-flow images at the original $1280{\times}2000$ resolution. No analogous filter is applied to the other domains, which do not exhibit this asset-list inflation.
\subsection{Benchmark Task Selection}
\label{app:task_selection}

Table~\ref{tab:task_selection} records the task selection used in the main evaluation, grouped by MMG2Skill-Bench domain. MMG2Skill-GUI covers all 10 OSWorld application domains released with the source benchmark. The per-domain count is taken proportionally from upstream OSWorld so that operation-heavy domains such as \texttt{multi\_apps}, \texttt{os}, and \texttt{vs\_code} keep their relative weight. MMG2Skill-Game takes 10 tasks each from \texttt{mine\_block}, \texttt{craft\_item}, and \texttt{smelt\_item}, the three task families released with the MineStudio harness. A small number of MMG2Skill-Game \texttt{smelt\_item} task IDs still carry a legacy \texttt{craft\_item:} prefix from the upstream naming convention. We treat the task-family grouping as authoritative and list those tasks under \texttt{smelt\_item}. MMG2Skill-Strategy contains 60 RLCard hands drawn from Doudizhu and Mahjong tasks.

\section{Extended Results and Robustness Checks}
\subsection{Complete Result Tables}
\label{app:full_results}
\begin{table*}[!t]
\centering
\renewcommand{\arraystretch}{1.2}
\resizebox{\textwidth}{!}{%
\begin{tabular}{@{}llccccccc@{}}
\toprule
\multirow{2}{*}{\textbf{Domain}} & \multirow{2}{*}{\textbf{Model}} & \multirow{2}{*}{\textbf{Vanilla}}
 & \multicolumn{3}{c}{\textbf{MMG2Skill (HTML)}}
 & \multicolumn{3}{c}{\textbf{MMG2Skill (Screen)}} \\
\cmidrule(lr){4-6}\cmidrule(lr){7-9}
 & & & full-run & early-stop & oracle & full-run & early-stop & oracle \\
\midrule
\multirow{6}{*}{\shortstack[l]{GUI\\(\%)}}
  & Claude-Opus-4.6        & 62.22 & 67.22 & 64.75 & 69.72 & 74.72 & 72.25 & 77.22 \\
  & GPT-5.5                & 39.74 & 49.74 & 57.24 & 59.74 & 44.74 & 52.24 & 57.24 \\
  & Claude-Sonnet-4.6      & 49.76 & 59.99 & 59.76 & 77.22 & 59.72 & 62.24 & 69.74 \\
  & Kimi-K2.6              & 69.75 & 77.25 & 77.25 & 79.75 & 74.77 & 77.27 & 79.77 \\
  & Gemini-3.1-Pro-Preview & 19.97 & 39.99 & 35.00 & 44.99 & 27.50 & 37.50 & 42.49 \\
  & Qwen3.6-Plus           & 15.00 & 39.99 & 39.99 & 42.49 & 35.00 & 40.00 & 42.49 \\
\cmidrule(lr){1-9}
\multirow{6}{*}{\shortstack[l]{Game\\(\%)}}
  & Claude-Opus-4.6        & 66.67 & 66.67 & 83.33 & 90.00 & 76.67 & 83.33 & 83.33 \\
  & GPT-5.5                & 73.33 & 66.67 & 80.00 & 80.00 & 76.67 & 83.33 & 83.33 \\
  & Claude-Sonnet-4.6      & 56.67 & 50.00 & 73.33 & 76.67 & 46.67 & 70.00 & 76.67 \\
  & Kimi-K2.6              & 16.67 & 23.33 & 46.67 & 56.67 & 26.67 & 46.67 & 46.67 \\
  & Gemini-3.1-Pro-Preview & 33.33 & 53.33 & 66.67 & 73.33 & 40.00 & 60.00 & 73.33 \\
  & Qwen3.6-Plus           & 16.67 & 26.67 & 46.67 & 53.33 & 23.33 & 26.67 & 30.00 \\
\cmidrule(lr){1-9}
\multirow{6}{*}{\shortstack[l]{Strategy-DD\\(\%)}}
  & Claude-Opus-4.6        & 16.67 & 23.33 & 46.67 & 50.00 & 36.67 & 53.33 & 53.33 \\
  & GPT-5.5                & 30.00 & 36.67 & 46.67 & 46.67 & 40.00 & 56.67 & 56.67 \\
  & Claude-Sonnet-4.6      & 13.33 & 23.33 & 46.67 & 50.00 & 13.33 & 40.00 & 40.00 \\
  & Kimi-K2.6              & 30.00 & 13.33 & 36.67 & 50.00 & 33.33 & 46.67 & 50.00 \\
  & Gemini-3.1-Pro-Preview & 43.33 & 40.00 & 53.33 & 56.67 & 33.33 & 63.33 & 66.67 \\
  & Qwen3.6-Plus           & 36.67 & 30.00 & 60.00 & 63.33 & 20.00 & 53.33 & 53.33 \\
\cmidrule(lr){1-9}
\multirow{6}{*}{\shortstack[l]{Strategy-MJ\\(\%)}}
  & Claude-Opus-4.6        & 66.67 & 75.00 & 75.00 & 96.67 & 71.67 & 71.67 & 95.00 \\
  & GPT-5.5                & 68.33 & 73.33 & 83.33 & 95.00 & 73.33 & 75.00 & 95.00 \\
  & Claude-Sonnet-4.6      & 66.67 & 66.67 & 76.67 & 91.67 & 68.33 & 78.33 & 91.67 \\
  & Kimi-K2.6              & 63.33 & 61.67 & 68.33 & 90.00 & 66.67 & 76.67 & 93.33 \\
  & Gemini-3.1-Pro-Preview & 65.00 & 65.00 & 68.33 & 85.00 & 68.33 & 71.67 & 90.00 \\
  & Qwen3.6-Plus           & 60.00 & 63.33 & 78.33 & 86.67 & 65.00 & 71.67 & 91.67 \\
\bottomrule
\end{tabular}%
}
\caption{\textbf{Complete per-model MMG2Skill-Bench results.} Success rates (\%) are shown for vanilla and MMG2Skill with HTML or screen observations under full-run, early-stop, and oracle variants within $N{=}5$. Strategy-DD and Strategy-MJ denote Doudizhu and Mahjong.}
\label{tab:appendix_full_results}
\end{table*}

Table~\ref{tab:appendix_full_results} extends the main-text results with the full-run variant of MMG2Skill, which reports the final attempt under $N{=}5$, and an \emph{oracle} upper bound that selects the earliest attempt with the highest score within the same $N{=}5$ budget. The oracle column upper-bounds what any analyzer-driven stopping rule could achieve given the same attempts.
On the four success-inferable domain rows (GUI, Game, Strategy-DD, Strategy-MJ), early-stop is no worse than the full-run variant in nearly all cells (early-stop $\ge$ full-run, 21/24 for HTML and 23/24 for Screen). This supports using early-stop as the deployable MMG2Skill view when the analyzer signal is calibrated.
The gap is especially pronounced on Game and Strategy-DD, where full-run scores drop substantially below early-stop, indicating late-stage regression that the analyzer can avoid when its success signal is calibrated.

The oracle column quantifies how much the analyzer leaves on the table relative to a hypothetically perfect stopping rule with the same attempt budget. In the HTML setting reported in the main table, the mean cell-level gap (oracle\,$-$\,early-stop) is $6.6$ points on GUI, $5.6$ on Game, and $4.4$ on Strategy-DD, within $8$--$11\%$ relative of the oracle ceiling. The gap is somewhat larger on Strategy-MJ ($+15.8$ pp, ${\approx}17\%$ relative), where the analyzer must infer success from complex hand-composition patterns and public game events rather than from a clean binary completion signal. The three-outcome engine payoff further complicates best-attempt selection, but the lower precision in Table~\ref{tab:analyzer_signal_calibration} shows that the difficulty is not merely a conservative stopping bias. Across both HTML and Screen, early-stop is already exactly tied with the oracle in $9$ of the $48$ cells. Together with the full-run comparison above, this places analyzer-based early stopping close to the ideal deployable rule on success-inferable tasks.

\subsection{Statistical Significance}
\label{app:significance}
\begin{table}[t]
\centering
\renewcommand{\arraystretch}{1.2}
\begin{tabular}{@{}llcccc@{}}
\toprule
\textbf{Domain} & \textbf{Modality} & $n_\text{task}$ & $\overline{\Delta}$ (\%) & Wilcoxon $p$ & Sign $p$ \\
\midrule
\multirow{2}{*}{GUI}
  & HTML   & 40 & $+12.9$ & $8.9\times10^{-4}$ & $2.0\times10^{-3}$ \\
  & Screen & 40 & $+14.2$ & $1.4\times10^{-4}$ & $1.0\times10^{-3}$ \\
\midrule
\multirow{2}{*}{Game}
  & HTML   & 30 & $+22.2$ & $2.4\times10^{-4}$ & $9.7\times10^{-6}$ \\
  & Screen & 30 & $+17.8$ & $6.5\times10^{-5}$ & $5.2\times10^{-6}$ \\
\midrule
\multirow{2}{*}{Strategy}
  & HTML   & 60 & $+15.0$ & $3.7\times10^{-7}$ & $4.7\times10^{-7}$ \\
  & Screen & 60 & $+16.5$ & $7.0\times10^{-8}$ & $7.4\times10^{-6}$ \\
\bottomrule
\end{tabular}%
\caption{\textbf{Per-task significance.} For each (domain, modality) column we average the per-task improvement $(\text{MMG2Skill}-\text{Vanilla})$ across the six evaluated models, then run a one-sided paired Wilcoxon signed-rank test and a one-sided sign test on the resulting per-task sequence.}
\label{tab:significance_task}
\end{table}

\begin{table}[t]
\centering
\renewcommand{\arraystretch}{1.2}
\begin{tabular}{@{}llccc@{}}
\toprule
\textbf{Domain} & \textbf{Modality} & $k$ & Stouffer $Z$ & Stouffer $p$ \\
\midrule
\multirow{2}{*}{GUI}
  & HTML   & 6 & $+4.27$ & $9.6\times10^{-6}$ \\
  & Screen & 6 & $+4.63$ & $1.8\times10^{-6}$ \\
\midrule
\multirow{2}{*}{Game}
  & HTML   & 5 & $+5.12$ & $1.5\times10^{-7}$ \\
  & Screen & 5 & $+4.61$ & $2.1\times10^{-6}$ \\
\midrule
\multirow{2}{*}{Strategy}
  & HTML   & 6 & $+5.74$ & $4.7\times10^{-9}$ \\
  & Screen & 6 & $+6.69$ & $1.1\times10^{-11}$ \\
\bottomrule
\end{tabular}%
\caption{\textbf{Stouffer-combined significance.} For each (domain, modality) column we run one paired Wilcoxon signed-rank test per model (concatenating hands from both underlying games within a model for Strategy), then combine the resulting one-sided $p$-values with Stouffer's $Z$-method.}
\label{tab:significance_stouffer}
\end{table}

Two complementary aggregations of the per-task scores confirm that MMG2Skill yields a statistically significant improvement over the Vanilla baseline on every (domain, modality) cell of Table~\ref{tab:main_results_long}. Scores are paired by task identifier within each underlying subset. All tests are one-sided ($\text{MMG2Skill}>\text{Vanilla}$) with continuity correction, and pairs with zero difference are excluded from the ranking per Wilcoxon's original procedure. This convention biases the resulting $p$ toward the null rather than inflating it.

The model-averaged per-task test shows that MMG2Skill helps the typical task in every domain. For each task we average $(\text{MMG2Skill}-\text{Vanilla})$ across the six models, then test whether the resulting per-task sequence exceeds zero with a paired Wilcoxon and a sign test. For Strategy, the per-task sequence is the concatenation of the per-task average $\Delta$ over its two underlying RLCard games (Doudizhu and Mahjong), giving 60 tasks. All six (domain, modality) cells reach Wilcoxon $p\leq 8.9\times10^{-4}$ and sign $p\leq 2.0\times10^{-3}$ (Table~\ref{tab:significance_task}).

A second aggregation that does not first collapse over models reaches the same conclusion. We run one paired Wilcoxon per model (concatenating both underlying games within a model for Strategy), then combine the one-sided $p$-values via Stouffer's $Z$-method. Combining $p$-values sidesteps the i.i.d. issue a naively pooled-diff Wilcoxon would incur when the same task identifier repeats across models. Every (domain, modality) cell yields combined $Z\geq +4.27$ and $p\leq 9.6\times10^{-6}$ (Table~\ref{tab:significance_stouffer}). The Wilcoxon and Stouffer results also remain significant under a Bonferroni correction over the six domain--modality cells. The sign test is reported as a distribution-free directional sanity check rather than the primary corrected test.

\subsection{Multi-Run Stability}
\label{app:run_stability}
\providecommand{\TODO}[1]{\textcolor{red}{[TODO: #1]}}
\begin{table}[t]
\centering
\begin{tabular}{@{}lllccc@{}}
\toprule
\textbf{Model} & \textbf{Method} & \textbf{Run} & \textbf{GUI} (\%) & \textbf{Game} (\%) & \textbf{Strategy} (\%) \\
\midrule
\multirow{9}{*}{Claude-Opus-4.6}
  & \multirow{3}{*}{Vanilla}
    & Run 1      & 62.22        & 66.67        & 41.67        \\
  & & Run 2      & 64.72   & 70.00   & 38.33   \\
  & & $|\Delta|$ & 2.50    & 3.33    & 3.34    \\
\cmidrule(lr){2-6}
  & \multirow{3}{*}{Raw Guide}
    & Run 1      & 59.75        & 66.67        & 44.17        \\
  & & Run 2      & 62.25   & 66.67   & 44.17   \\
  & & $|\Delta|$ & 2.50    & 0.00    & 0.00    \\
\cmidrule(lr){2-6}
  & \multirow{3}{*}{MMG2Skill}
    & Run 1      & 64.75        & 83.33        & 60.83        \\
  & & Run 2      & 70.00   & 83.33   & 62.50   \\
  & & $|\Delta|$ & 5.25    & 0.00    & 1.67    \\
\midrule
\multirow{9}{*}{GPT-5.5}
  & \multirow{3}{*}{Vanilla}
    & Run 1      & 39.74        & 73.33        & 49.17        \\
  & & Run 2      & 34.74   & 73.33   & 46.67   \\
  & & $|\Delta|$ & 5.00    & 0.00    & 2.50    \\
\cmidrule(lr){2-6}
  & \multirow{3}{*}{Raw Guide}
    & Run 1      & 37.24        & 76.67        & 50.00        \\
  & & Run 2      & 37.24   & 80.00   & 48.33   \\
  & & $|\Delta|$ & 0.00    & 3.33    & 1.67    \\
\cmidrule(lr){2-6}
  & \multirow{3}{*}{MMG2Skill}
    & Run 1      & 57.24        & 80.00        & 65.00        \\
  & & Run 2      & 52.24   & 86.67   & 69.17   \\
  & & $|\Delta|$ & 5.00    & 6.67    & 4.17    \\
\midrule
\multirow{9}{*}{Kimi-K2.6}
  & \multirow{3}{*}{Vanilla}
    & Run 1      & 69.75        & 16.67        & 46.67        \\
  & & Run 2      & 69.75   & 20.00   & 45.83   \\
  & & $|\Delta|$ & 0.00    & 3.33    & 0.84    \\
\cmidrule(lr){2-6}
  & \multirow{3}{*}{Raw Guide}
    & Run 1      & 74.75        & 16.67        & 40.83        \\
  & & Run 2      & 72.25   & 16.67   & 43.33   \\
  & & $|\Delta|$ & 2.50    & 0.00    & 2.50    \\
\cmidrule(lr){2-6}
  & \multirow{3}{*}{MMG2Skill}
    & Run 1      & 77.25        & 46.67        & 52.50        \\
  & & Run 2      & 77.25   & 46.67   & 50.83   \\
  & & $|\Delta|$ & 0.00    & 0.00    & 1.67    \\
\bottomrule
\end{tabular}%
\caption{\textbf{Multi-run stability on the three MMG2Skill-Bench domains (HTML modality).} Run~1 reproduces Table~\ref{tab:main_results_long}; Run~2 is an independent rerun with the same hyperparameters and attempt budget ($N{=}5$). $|\Delta|$ reports the absolute difference between runs.}
\label{tab:appendix_run_stability}
\end{table}

To check that the main-table cells are not artefacts of a single rollout, we re-run all three methods (Vanilla, Raw Guide, MMG2Skill) on the three MMG2Skill-Bench domains for three representative backbones that span the per-model score range in Table~\ref{tab:main_results_long}: Claude-Opus-4.6, GPT-5.5, and Kimi-K2.6. Each rerun uses the same hyperparameters and attempt budget ($N{=}5$) as the main experiment. Table~\ref{tab:appendix_run_stability} reports the per-cell scores for both runs together with the absolute per-cell difference $|\Delta|$, so that the MMG2Skill\,$-$\,Vanilla gap can be compared across runs rather than only the MMG2Skill score in isolation.

Across the $27$ (model, method, domain) cells, the mean absolute run-to-run difference is $2.14$ percentage points, roughly one task or hand for these domains. This level of variation does not affect the main conclusion: the macro-averaged MMG2Skill$-$Vanilla gap remains positive for all three representative backbones in both runs.

\section{Revision and Analyzer Diagnostics}
\subsection{Per-Model Revision Dynamics}
\label{app:per_model_revision}

\begin{figure}[t]
\centering
\includegraphics[width=\linewidth]{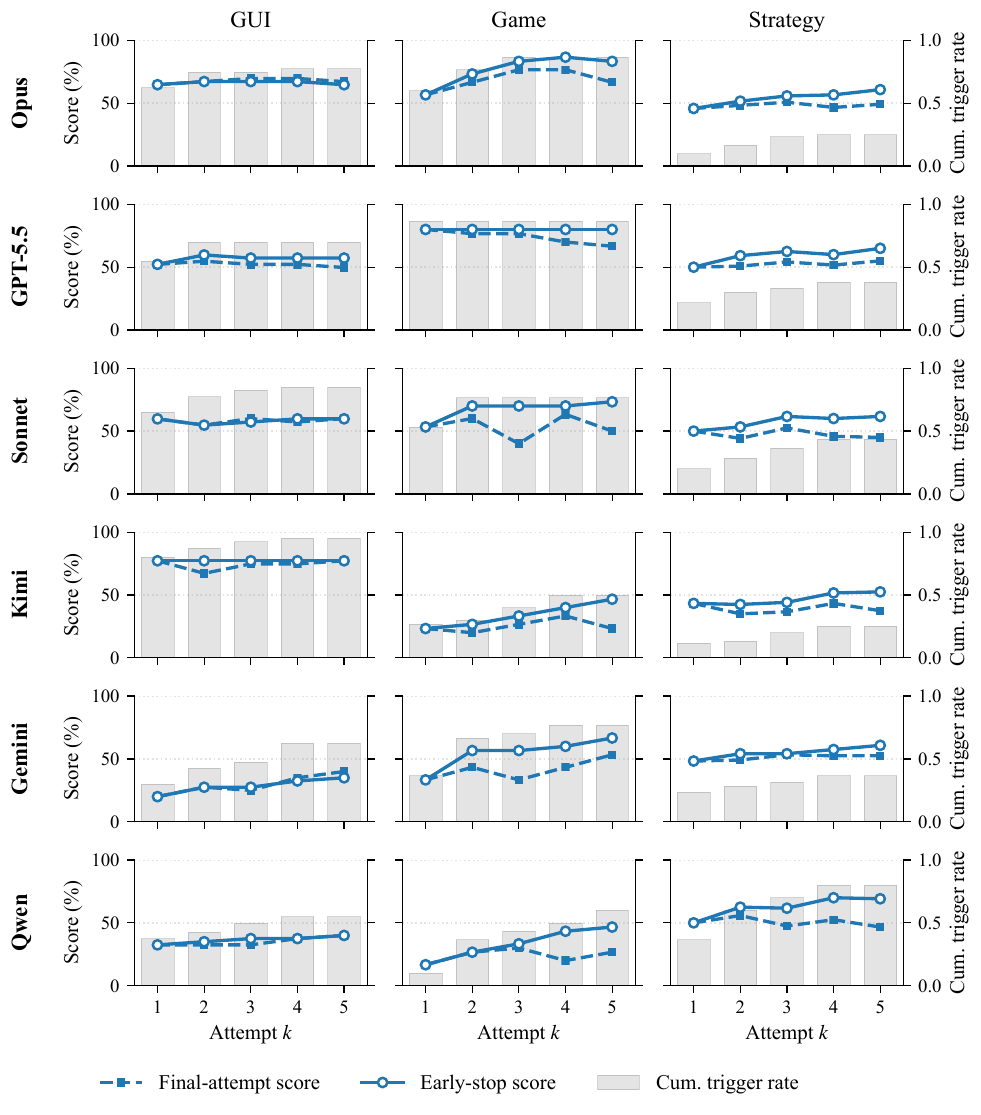}
\caption{\textbf{Per-(model, domain) revision dynamics.} This expanded view of Figure~\ref{fig:revision_dynamics} shows one panel per (model, domain) pair. Each panel reports absolute mean full-run score (dashed, filled squares), early-stop score (solid, open circles), and the cumulative \texttt{likely\_success} trigger rate as a faded right-axis bar.}
\label{fig:per_model_revision_dynamics}
\end{figure}

This section extends the main-text revision dynamics (Figure~\ref{fig:revision_dynamics}) by showing one panel per (model, domain) pair. Within-domain variance is substantial. Some models plateau immediately, while others show steady improvement through all five attempts. The same per-model traces also expose cases where the full-run view regresses at late budgets, most prominently in Game for Sonnet and in Strategy for Qwen, which motivates the early-stop ablation in \S\ref{sec:exp:early_stop}.

\subsection{Extended Revision Horizon}
\label{app:extended_horizon}

\begin{figure}[t]
\centering
\includegraphics[width=\linewidth]{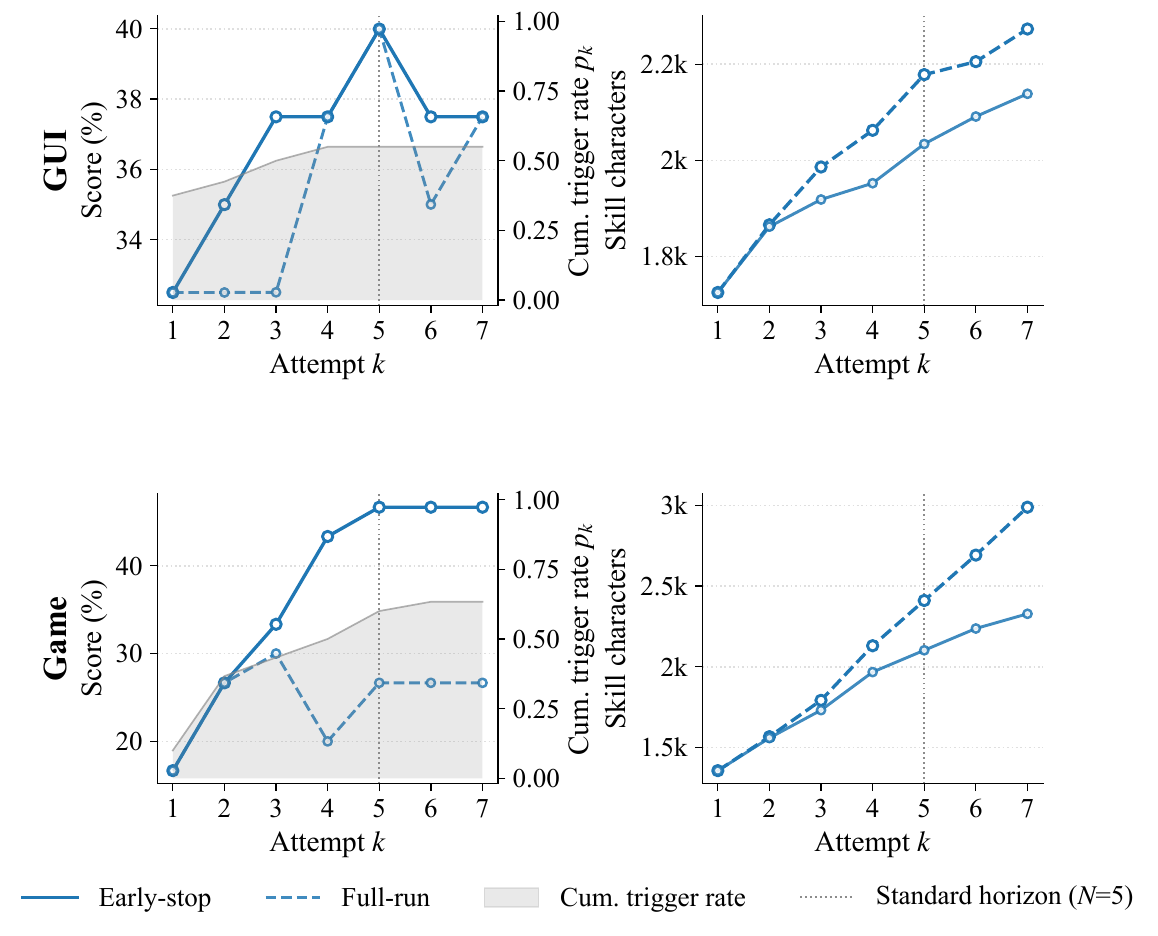}
\caption{\textbf{Extended revision horizon} ($N{=}7$) on Qwen3.6-Plus for the GUI and Game domains. Panels report per-attempt mean score (\%) under the full-run and early-stop policies, the cumulative \texttt{likely\_success} trigger rate on the right axis, and skill character count.}
\label{fig:extended_horizon}
\end{figure}

We extend the revision horizon to $N{=}7$ on Qwen3.6-Plus for the GUI and Game domains (Figure~\ref{fig:extended_horizon}). The early-stop score plateaus by $N{=}5$ on both domains while the full-run curve continues to oscillate, so the $N{=}5$ budget used in the main text captures the available revision benefit. Extending to $N{=}7$ yields no further improvement under early stopping, and reinforces the non-monotonicity flagged in \S\ref{sec:exp:revision} for the full-run view.

\subsection{Skill-Size Dynamics across Attempts}
\label{app:skill_chars}

\begin{figure*}[t]
\centering
\includegraphics[width=\linewidth]{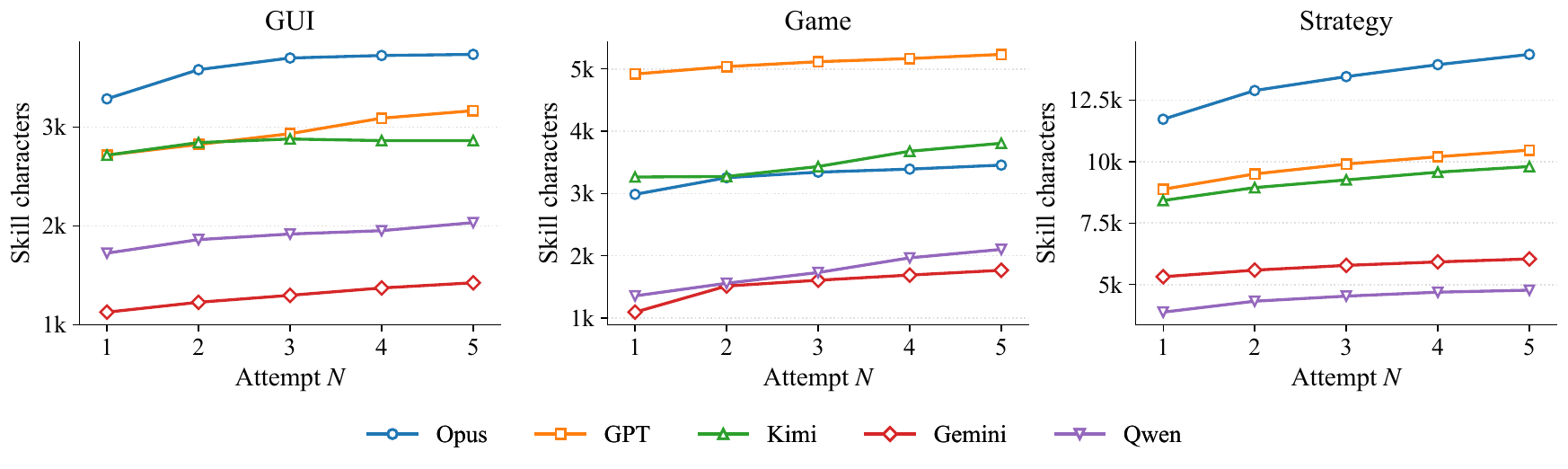}
\caption{\textbf{Early-stop skill character count across attempts $N$} under MMG2Skill on the three MMG2Skill-Bench domains. Each curve reports one backbone's task-mean skill-set size selected by the \texttt{likely\_success} early-stop rule within budget $N$. Bands show standard deviation.}
\label{fig:skill_convergence}
\end{figure*}

Figure~\ref{fig:skill_convergence} reports the task-mean character count of the early-stop skill set at $N{=}1,\ldots,5$ across five backbones. Skill size increases with the available revision budget but the growth visibly flattens after the first attempts, indicating that the revision loop adds most reusable procedural content early. Because the plotted skill set is selected by the \texttt{likely\_success} stopping rule, the same mechanism that saves environment attempts also bounds the deployed skill prompt size.

\subsection{Direct Root-Cause Prompting Diagnostic}
\label{app:raw_guide_root_cause}

\begin{figure}[t]
\centering
\includegraphics[width=\linewidth]{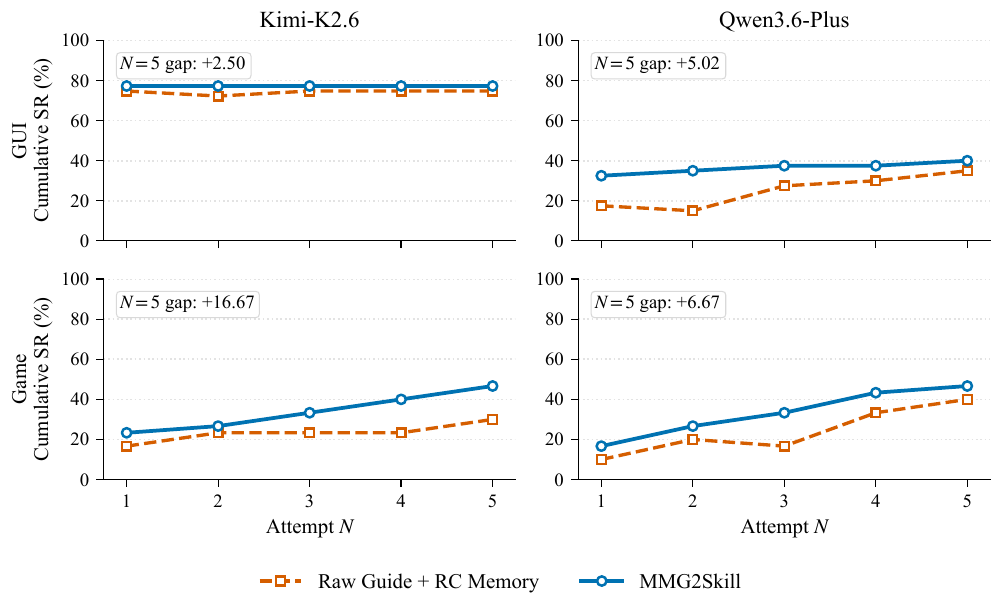}
\caption{\textbf{Direct root-cause prompting diagnostic.} Each panel reports cumulative success rate under the same early-stop policy used by MMG2Skill.}
\label{fig:raw_guide_root_cause_diagnostic}
\end{figure}

Figure~\ref{fig:raw_guide_root_cause_diagnostic} tests whether MMG2Skill's revision gain can be reduced to extra test-time attempts and root-cause memory in the prompt. The diagnostic variant appends previous analyzer root causes to the raw guide and uses the same early-stop policy as MMG2Skill, but it disables both skill construction and refinement. This gives the raw-guide baseline access to trajectory-level feedback without giving it an editable skill file. The diagnostic is scoped to MMG2Skill-GUI and MMG2Skill-Game, where the main results show the largest revision-driven gains.

The diagnostic variant improves in some settings, but it remains below MMG2Skill at $N{=}5$ in all four model--domain pairs. The gaps are largest on MMG2Skill-Game, where failures often require concrete updates to state checks, crafting order, or recovery behavior. These results support the interpretation in \S\ref{sec:exp:ablation}. Analyzer feedback is not merely useful as additional context. It becomes more effective when the refiner turns it into persistent edits to the execution interface.

\subsection{Analyzer-Based Stopping Diagnostics}
\label{app:attempt_dynamics}

This appendix groups diagnostics for the analyzer signal that drives early stopping.

\subsubsection{Analyzer Signal Calibration}
\label{app:signal_calibration}

\begin{figure*}[htbp]
\centering
\includegraphics[width=\linewidth]{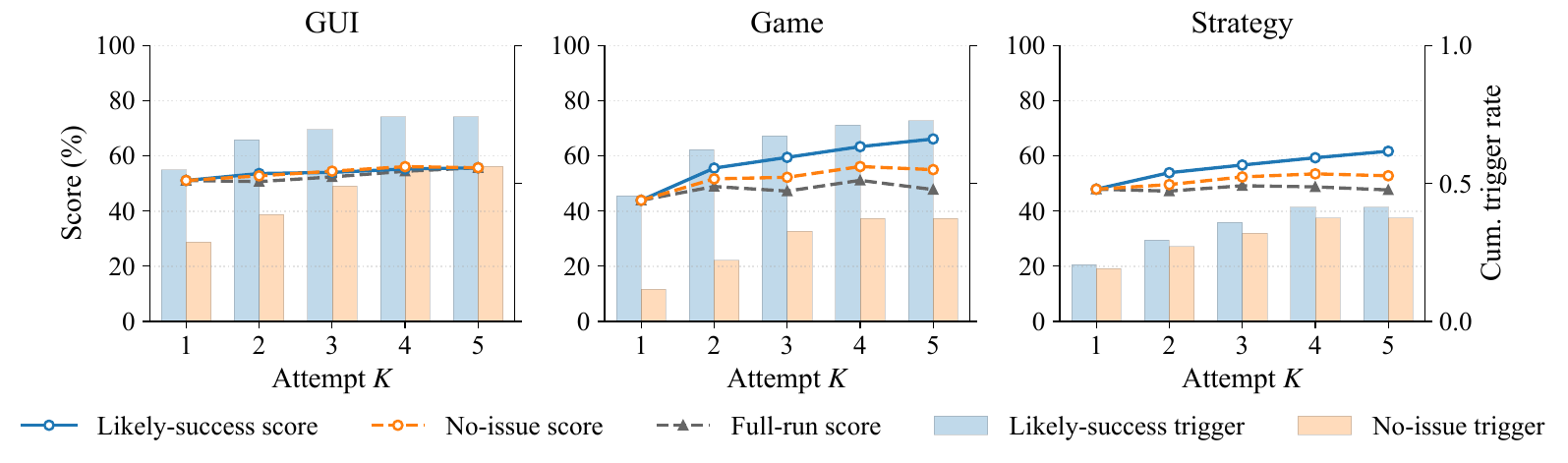}
\caption{\textbf{Analyzer signal calibration.} The figure compares \texttt{likely\_success} and \texttt{no\_issue} trigger rates, early-stop scores, and gaps to full-run across attempts $K$. }
\label{fig:signal_calibration}
\end{figure*}

\begin{table*}[!t]
\centering
\renewcommand{\arraystretch}{1.15}
\resizebox{\textwidth}{!}{%
\begin{tabular}{@{}llcccccccccc@{}}
\toprule
\multirow{2}{*}{\textbf{Model}} & \multirow{2}{*}{\textbf{Signal}} & \multicolumn{2}{c}{\textbf{GUI}} & \multicolumn{2}{c}{\textbf{Game}} & \multicolumn{2}{c}{\textbf{Strategy-DD}} & \multicolumn{2}{c}{\textbf{Strategy-MJ}} & \multicolumn{2}{c}{\textbf{Hold'em}} \\
\cmidrule(lr){3-4} \cmidrule(lr){5-6} \cmidrule(lr){7-8} \cmidrule(lr){9-10} \cmidrule(lr){11-12}
 & & \textbf{P (\%)} & \textbf{R (\%)} & \textbf{P (\%)} & \textbf{R (\%)} & \textbf{P (\%)} & \textbf{R (\%)} & \textbf{P (\%)} & \textbf{R (\%)} & \textbf{P (\%)} & \textbf{R (\%)} \\
\midrule
\multirow{2}{*}{Claude-Opus-4.6}
  & LS & 83.9  & 100.0 & 96.2  & 100.0 & 100.0 & 100.0 & 0.0   & 0.0   & 27.3  & 28.1 \\
  & NI & 88.9  & 92.3  & 100.0 & 82.6  & 100.0 & 88.9  & ---   & 0.0   & 14.8  & 12.5 \\
\cmidrule(lr){1-12}
\multirow{2}{*}{GPT-5.5}
  & LS & 82.1  & 100.0 & 92.3  & 100.0 & 100.0 & 100.0 & 100.0 & 45.0  & 7.1   & 25.9 \\
  & NI & 90.0  & 85.7  & 81.8  & 40.9  & 100.0 & 92.3  & 75.0  & 40.0  & 8.1   & 28.0 \\
\cmidrule(lr){1-12}
\multirow{2}{*}{Claude-Sonnet-4.6}
  & LS & 70.6  & 100.0 & 91.3  & 95.5  & 100.0 & 92.9  & 76.9  & 62.5  & 11.9  & 66.7 \\
  & NI & 74.1  & 80.0  & 100.0 & 38.9  & 100.0 & 62.5  & 80.0  & 33.3  & 5.2   & 22.2 \\
\cmidrule(lr){1-12}
\multirow{2}{*}{Kimi-K2.6}
  & LS & 81.6  & 100.0 & 80.0  & 85.7  & 100.0 & 72.7  & 71.4  & 41.7  & 4.2   & 50.0 \\
  & NI & 85.7  & 77.4  & 100.0 & 37.5  & ---   & 0.0   & ---   & 0.0   & 0.0   & 0.0 \\
\cmidrule(lr){1-12}
\multirow{2}{*}{\shortstack[l]{Gemini-3.1-\\Pro-Preview}}
  & LS & 52.0  & 92.9  & 78.3  & 90.0  & 82.3  & 87.5  & 80.0  & 36.4  & 8.7   & 100.0 \\
  & NI & 70.6  & 75.0  & 76.2  & 80.0  & 58.3  & 100.0 & 33.3  & 100.0 & 9.3   & 100.0 \\
\cmidrule(lr){1-12}
\multirow{2}{*}{Qwen3.6-Plus}
  & LS & 68.2  & 93.8  & 77.8  & 100.0 & 90.0  & 100.0 & 60.7  & 100.0 & 12.0  & 100.0 \\
  & NI & 81.2  & 86.7  & 66.7  & 50.0  & 80.0  & 100.0 & 57.7  & 100.0 & 13.7  & 90.9 \\
\midrule
\multirow{2}{*}{\textit{Aggregate}}
  & LS & \textbf{74.2} & \textbf{98.5} & \textbf{87.0} & \textbf{95.8} & \textbf{94.2} & \textbf{93.1} & \textbf{71.4} & \textbf{49.5} & \textbf{9.9}  & \textbf{54.5} \\
  & NI & \textbf{82.2} & \textbf{82.8} & \textbf{86.6} & \textbf{58.6} & \textbf{79.7} & \textbf{85.9} & \textbf{51.5} & \textbf{45.3} & \textbf{8.7}  & \textbf{41.2} \\
\bottomrule
\end{tabular}%
}
\caption{\textbf{Per-domain early-stop decision calibration on the three MMG2Skill-Bench domains and the Hold'em diagnostic.} Calibration is reported for \texttt{likely\_success} (LS) vs.\ \texttt{no\_issue} (NI) under the HTML guide variant. A signal is positive if it triggers within $N{=}5$; ground truth is positive when the resulting score is $>0$. Cells with no triggered tasks (and thus undefined precision) are shown as ``---''.}
\label{tab:analyzer_signal_calibration}
\end{table*}

Besides the main \texttt{likely\_success} outcome assessment, we also evaluate a stricter \texttt{no\_issue} signal that fires only when the analyzer reports no remaining execution issue in the trajectory. Table~\ref{tab:analyzer_signal_calibration} and Figure~\ref{fig:signal_calibration} give the full calibration breakdown for the two candidate stopping signals. We define predicted positive as \texttt{likely\_success}, predicted negative as \texttt{uncertain} or \texttt{likely\_failure}, and ground truth as the offline benchmark grader after benchmark-specific binarization (score $>0$). A ``---'' entry indicates that no predictions of that class were made.

The detailed numbers support the main-text choice of \texttt{likely\_success}. It has the best precision--recall balance across the success-inferable domains, while \texttt{no\_issue} usually trades too much recall for small precision gains.

Reading the table across domains traces a clear degradation gradient as task outcomes become harder to infer from the trajectory. GUI, Game, and Strategy-DD all carry a public completion signal, and aggregate \texttt{likely\_success} on these three domains keeps both precision and recall high ($\ge$87/93). Strategy-MJ is the first regime where the two metrics drop together, to $71.4\%$ precision and $49.5\%$ recall, for the hand-composition and three-outcome reasons already noted in \S\ref{app:full_results}. The drop is symmetric rather than a conservative shift, which already hints at a harder regime once trajectory-level inference is removed entirely. We isolate that regime next on No-Limit Hold'em.

\subsubsection{No-Limit Hold'em: Private-Information Boundary}
\label{app:holdem_boundary}

No-Limit Hold'em runs on the same RLCard engine as MMG2Skill-Strategy but is held out of the MMG2Skill-Strategy task pool. The boundary is drawn by the success-inferability criterion stated in \S\ref{sec:benchmark:tasks}. A task qualifies when its outcome can be inferred from the agent-visible trajectory or the public final state, even when the environment hides some state from the agent. Doudizhu and Mahjong satisfy this condition because each hand resolves through public events such as cards played out, a declared \emph{hu}, or an exhausted wall, none of which require reading opponent hands from the trajectory. No-Limit Hold'em fails it because chip payoff depends on hole cards that are revealed only at showdown, and many hands end before showdown through betting alone. The trajectory can therefore expose decision quality but cannot reliably reveal whether the hand was won, which is the variable that any analyzer-based stopping signal would need to track.

\begin{table}[t]
\centering
\renewcommand{\arraystretch}{1.15}
\begin{tabular}{@{}lc@{}}
\toprule
\textbf{Quantity} & \textbf{Value} \\
\midrule
Vanilla score & 2.44 \\
MMG2Skill early-stop score & 3.59 \\
MMG2Skill full-run score & 2.23 \\
\texttt{likely\_success} precision & 9.9\% \\
\texttt{likely\_success} recall & 54.5\% \\
\texttt{likely\_success} trigger rate & 74.2\% \\
Actual positive rate & 13.4\% \\
Mean stopping attempt $\bar{k}$ & 2.42 \\
\bottomrule
\end{tabular}
\caption{\textbf{No-Limit Hold'em private-information boundary for analyzer-based stopping.} Model-mean chip payoffs under HTML guides; payoffs depend on opponent private cards absent from the agent-visible trajectory.}
\label{tab:holdem_boundary_case}
\end{table}

Table~\ref{tab:holdem_boundary_case} quantifies the resulting miscalibration. The \texttt{likely\_success} signal reaches only $9.9\%$ precision, triggering on $74.2\%$ of tasks despite an actual positive rate of $13.4\%$. Table~\ref{tab:holdem_full_results} reports the corresponding per-model full-run, early-stop, and oracle results. The oracle gap is much larger than on success-inferable tasks because a trajectory-only analyzer cannot observe the private cards that determine chip payoff. The higher early-stop score than full-run score should therefore not be read as a reliable stopping-policy result. In private-information settings where the outcome is not inferable from the trajectory, analyzer-based stopping should be disabled or replaced with external outcome feedback. The skill-revision loop itself can still be run.

\begin{table*}[t]
\centering
\renewcommand{\arraystretch}{1.2}
\resizebox{\textwidth}{!}{%
\begin{tabular}{@{}lccccccc@{}}
\toprule
\multirow{2}{*}{\textbf{Model}} & \multirow{2}{*}{\textbf{Vanilla}}
 & \multicolumn{3}{c}{\textbf{MMG2Skill (HTML)}}
 & \multicolumn{3}{c}{\textbf{MMG2Skill (Screen)}} \\
\cmidrule(lr){3-5}\cmidrule(lr){6-8}
 & & full-run & early-stop & oracle & full-run & early-stop & oracle \\
\midrule
Claude-Opus-4.6        &  4.89 &  5.53 &  5.83 & 11.28 &  4.83 &  4.73 &  8.53 \\
GPT-5.5                &  0.79 &  3.57 &  4.26 &  9.93 &  2.39 &  2.33 &  7.41 \\
Claude-Sonnet-4.6      &  3.71 &  2.22 &  3.69 &  8.45 &  2.93 &  5.01 &  9.20 \\
Kimi-K2.6              &  1.40 & $-$1.31 &  2.25 &  3.78 &  3.29 &  3.47 &  5.49 \\
Gemini-3.1-Pro-Preview &  2.95 &  2.31 &  3.19 &  4.96 &  2.03 &  2.22 &  3.17 \\
Qwen3.6-Plus           &  0.90 &  1.09 &  2.32 &  4.76 &  1.64 & $-$0.79 &  9.00 \\
\bottomrule
\end{tabular}%
}
\caption{\textbf{Per-model results for the No-Limit Hold'em private-information diagnostic.} Full-run always uses the final attempt within $N{=}5$; early-stop terminates at the first \texttt{likely\_success} assessment within $N{=}5$; oracle picks the first positive-payoff attempt within $N{=}5$, otherwise the final attempt.}
\label{tab:holdem_full_results}
\end{table*}

\section{Efficiency and Deployment Cost}
\subsection{Step Efficiency on Step-Meaningful Domains}
\label{app:step_efficiency}

Table~\ref{tab:appendix_step_efficiency} reports displayed-attempt step counts for the GUI and Game domains, the two domains where step count reflects execution length rather than fixed turn progression. The attempt selection mirrors the main results.

Step count is a diagnostic rather than the main objective. Shorter rollouts are not always better, since failed attempts can also terminate early. On these two step-meaningful domains, however, the aggregate pattern is informative. Raw Guide is essentially flat relative to Vanilla at the macro level, while MMG2Skill reduces macro-average steps from $10.08$ to $9.21$ on the GUI domain and from $41.17$ to $36.08$ on the Game domain. The reduction supports the RQ1 interpretation that skills improve execution by grounding procedures into state cues and actionable next steps, making the agent's behavior more directed and reducing avoidable search.

\begin{table*}[t]
\centering
\renewcommand{\arraystretch}{1.15}
\resizebox{\textwidth}{!}{%
\begin{tabular}{@{}llrrrrr@{}}
\toprule
\textbf{Domain} & \textbf{Model} & \textbf{Vanilla} & \textbf{Raw Guide} & \textbf{MMG2Skill} & \textbf{Raw$-$Van} & \textbf{MMG$-$Van} \\
\midrule
\multirow{7}{*}{GUI}
  & Claude-Opus-4.6        &  8.60 &  9.12 &  9.05 & +0.52 & +0.45 \\
  & GPT-5.5                &  8.07 &  7.78 &  9.43 & -0.29 & +1.36 \\
  & Claude-Sonnet-4.6      &  9.35 &  9.07 &  7.20 & -0.28 & -2.15 \\
  & Kimi-K2.6              & 10.07 & 10.28 &  9.43 & +0.21 & -0.64 \\
  & Gemini-3.1-Pro-Preview & 11.85 & 11.72 &  9.38 & -0.13 & -2.47 \\
  & Qwen3.6-Plus           & 12.55 & 13.12 & 10.78 & +0.57 & -1.77 \\
  & \textbf{Macro-avg}     & \textbf{10.08} & \textbf{10.18} & \textbf{9.21} & \textbf{+0.10} & \textbf{-0.87} \\
\midrule
\multirow{7}{*}{Game}
  & Claude-Opus-4.6        & 37.57 & 37.20 & 31.27 & -0.37 & -6.30 \\
  & GPT-5.5                & 34.13 & 31.30 & 30.73 & -2.83 & -3.40 \\
  & Claude-Sonnet-4.6      & 41.20 & 41.67 & 35.97 & +0.47 & -5.23 \\
  & Kimi-K2.6              & 54.67 & 48.70 & 42.37 & -5.97 & -12.30 \\
  & Gemini-3.1-Pro-Preview & 33.00 & 36.80 & 30.13 & +3.80 & -2.87 \\
  & Qwen3.6-Plus           & 46.47 & 52.23 & 46.03 & +5.76 & -0.44 \\
  & \textbf{Macro-avg}     & \textbf{41.17} & \textbf{41.32} & \textbf{36.08} & \textbf{+0.15} & \textbf{-5.09} \\
\bottomrule
\end{tabular}%
}
\caption{\textbf{Step efficiency on step-meaningful domains.} Mean displayed-attempt steps under the main HTML setting on the GUI and Game domains, with attempt selection mirroring Table~\ref{tab:main_results_long}. The Strategy domain is omitted because hand length there is jointly determined by agent play, opponent play, and engine-side termination, so fewer steps can equally indicate a quick win or a quick loss.}
\label{tab:appendix_step_efficiency}
\end{table*}

\subsection{API-Call and Token Budget Decomposition}
\label{app:token_budget}

\begin{figure*}[t]
\centering
\includegraphics[width=\linewidth]{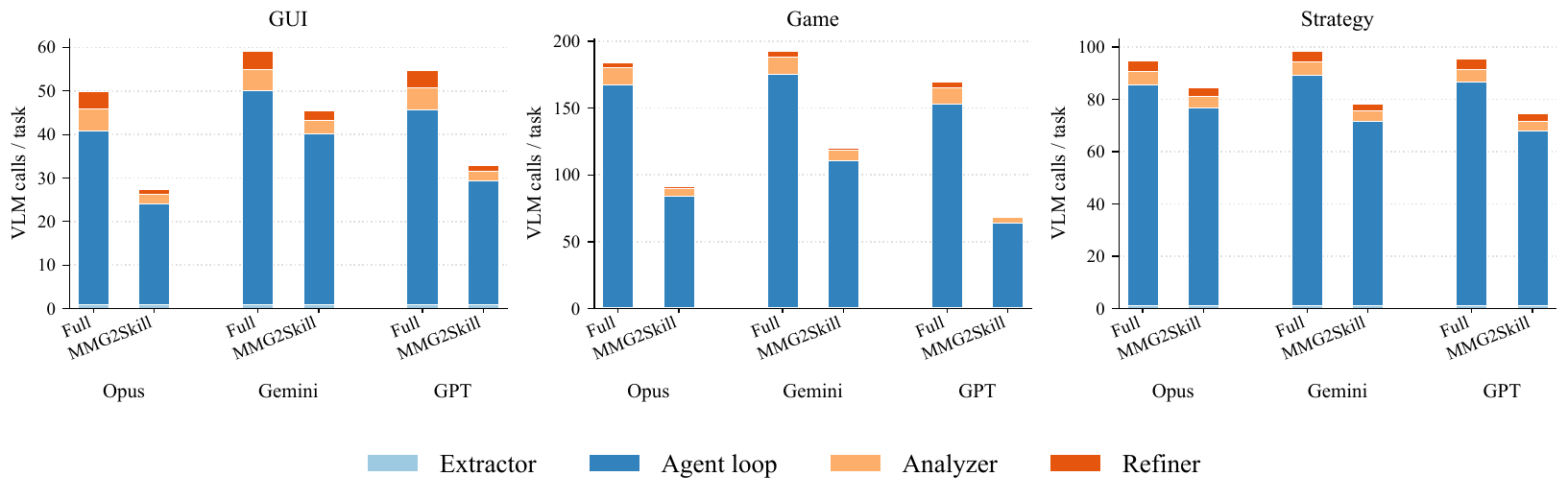}
\caption{\textbf{Per-task VLM API calls by pipeline stage} on the three MMG2Skill-Bench domains (Strategy averages Doudizhu and Mahjong, Hold'em excluded). Analyzer calls are estimated from the appendix chunk size $C$ and each attempt's decision steps as $\lceil L/C \rceil$. Refiner calls occur between counted attempts.}
\label{fig:api_call_stack}
\end{figure*}

Figure~\ref{fig:api_call_stack} decomposes online deployment cost at the API-call level. Agent-loop calls dominate every stack because each rollout step is a VLM action call. The reduction from early stopping is therefore driven mainly by fewer rollout attempts rather than by the one-shot extractor, whose cost is fixed at one call per task. The largest call savings appear on Game, where MMG2Skill reduces per-task calls by 37.72\% to 59.33\% across the three backbones. GUI saves 22.99\% to 45.06\%, while Strategy saves 10.95\% to 22.15\%, consistent with later stopping in Strategy. Refiner calls also fall from four under full-run to the average number of revisions made before the selected stopping attempt.

\begin{figure*}[!t]
\centering
\includegraphics[width=\linewidth]{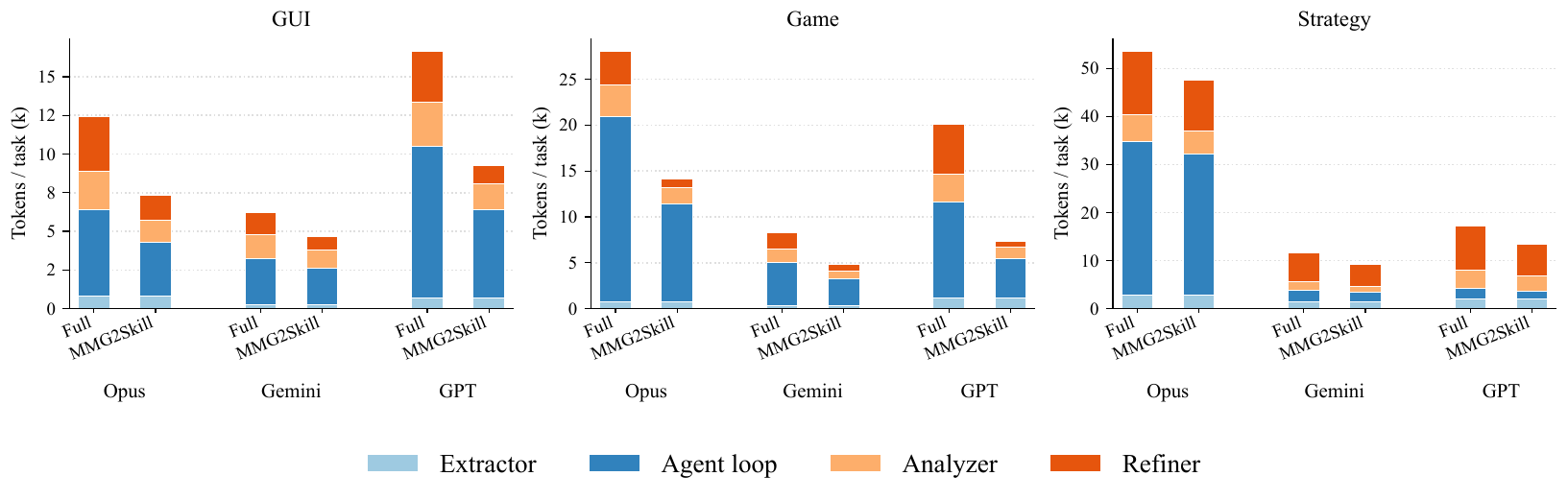}
\caption{\textbf{Per-task token consumption by pipeline stage} on the same three domains and backbones as Figure~\ref{fig:api_call_stack}, counted with the \texttt{o200k} tiktoken. \emph{Full} sums all $N{=}5$ attempts. \emph{MMG2Skill} stops at the first \texttt{likely\_success}.}
\label{fig:token_usage_stack}
\end{figure*}

Figure~\ref{fig:token_usage_stack} complements the call-count view with token consumption. Early stopping reduces the per-task token budget in every (domain, model) cell across the three backbones with full per-stage logs (Claude-Opus-4.6, Gemini-3.1-Pro-Preview, GPT-5.5), with larger savings on GUI and Game than on Strategy, matching the cumulative \texttt{likely\_success} trigger-rate profile in Figure~\ref{fig:signal_calibration}. Agent loop is the heaviest stage on GUI and Game, while on Strategy it is overtaken by Refiner for Gemini-3.1-Pro-Preview and GPT-5.5, with Claude-Opus-4.6 remaining loop-dominated.

\section{Case Study}
\label{app:case_study}

The four cases below ground the qualitative claims in \S\ref{sec:exp:main}, \S\ref{sec:exp:revision}, and \S\ref{sec:exp:early_stop} into concrete skill-diff and trajectory evidence. Each case uses the same template. It reports the MMG2Skill-Bench domain, task, and backbone, identifies the runtime-vs-guide grounding gap, shows the relevant skill or trajectory snapshot, and records the final outcome.

\subsection{MMG2Skill-GUI: GIMP Crop Skill Refinement}
\label{app:case_gimp_crop}

\noindent\textbf{Revision mechanism.} This case instantiates the runtime-contract mismatch discussed in \S\ref{sec:exp:revision}. The first attempt scores $0$ because the agent declares \texttt{DONE} after issuing the GIMP batch command even though \texttt{cropped.png} is not produced. The refiner localizes the fix to \texttt{run-gimp-crop-command/SKILL.md}, lengthening the wait specification and adding an explicit \texttt{ls -la cropped.png} verification gate before completion. The next attempt scores $1.0$, showing a revision benefit that static extraction cannot provide.

\begin{instructionbox}
\textbf{Task instruction.} Launch GIMP from the command line to edit ``cola.png'' and crop the top 20\% off the image for my avatar as ``cropped.png''.
\end{instructionbox}

\begin{diffbox}{Refiner edit: \texttt{run-gimp-crop-command/SKILL.md}\\[-1pt]\scriptsize \texttt{multi\_apps/91190194}; \texttt{claude-sonnet-4.6}; score $0 \to 1.0$}
\begin{Verbatim}[commandchars=\?\{\},fontsize=\scriptsize,breaklines=true,breaksymbolleft={},breaksymbolright={}]
 ---
 name: run-gimp-crop-command
 description: Launches GIMP in non-interactive batch mode to crop the top 20% from "cola.png" and save the result as "cropped.png".
 images:
   - /path/to/GimpCrop-01.png
 ---

 ## Steps
 1. Open a terminal by pressing `pyautogui.hotkey('ctrl', 'alt', 't')`
 2. Navigate to the directory containing `cola.png` (adjust path as needed). For example, if it is in the home directory:
    ```
    pyautogui.typewrite('cd ~', interval=0.05)
    pyautogui.press('enter')
    ```
 3. Run GIMP in headless batch mode with the crop script, typing the full command and pressing Enter:
    ```
    pyautogui.typewrite('gimp -i -b \'(crop-top-percent "cola.png" "cropped.png" 0.20)\' -b \'(gimp-quit 0)\'', interval=0.05)
    pyautogui.press('enter')
    ```
    ![Terminal showing gimp batch command executing successfully](GimpCrop-01.png)
?textcolor{diffrem}{- 4. Wait for GIMP to finish processing -- the terminal prompt will return once complete.}
?textcolor{diffadd}{+ 4. Wait for GIMP to finish -- this can take 2--4 minutes. Keep watching the terminal until a new shell prompt appears below the gimp command output before proceeding.}
?textcolor{diffadd}{+ 5. Once the shell prompt has returned, verify the output file was created by running:}
?textcolor{diffadd}{+    ```}
?textcolor{diffadd}{+    pyautogui.typewrite('ls -la cropped.png', interval=0.05)}
?textcolor{diffadd}{+    pyautogui.press('enter')}
?textcolor{diffadd}{+    ```}
?textcolor{diffadd}{+    Confirm the file exists and has a non-zero size in the terminal output before declaring the task complete.}
 ## Expected Result
 A new file `cropped.png` is created in the same directory as `cola.png`, containing the original image with the top 20% of pixels removed (i.e., the image height is 80% of the original, starting from 20% down).
\end{Verbatim}
\end{diffbox}

\subsection{MMG2Skill-GUI: Essay Submission Packaging}
\label{app:case_essay_submission}

\noindent\textbf{False-positive root cause.} This case gives the concrete false-positive pattern referenced in \S\ref{sec:exp:early_stop}. The rollout creates an archive and shows no visible execution error, so the analyzer fires \texttt{likely\_success}. The oracle still assigns score $0$ because the observed trajectory does not verify all submission requirements.

\begin{instructionbox}
\textbf{Task instruction.} The guidelines for submitting our essay work are provided in the ``reminder.docx'' on the Desktop. My essay is stored as a docx file in /home/user. Please assist me in organizing the files for submission according to the specified requirements.
\end{instructionbox}

For readability, we abbreviate the essay filename as \texttt{Recruitment\_...\_Europe.docx} and the converted PDF as \texttt{Recruitment\_...\_Europe.pdf} in the trajectory excerpts below.

\begin{failbox}{MMG2Skill Agent Trajectory (\texttt{multi\_apps/8df7e444}, \texttt{claude-sonnet-4.6})}
===== Step 1 (predict_num=1, phase=act) =====
I can see the requirements: export essay as PDF, pack the writer file along with the pdf into a ZIP named 'essay_submission.zip'.

<action>
import subprocess, os

# Find the essay docx
result = subprocess.run(['ls', '/home/user/'], capture_output=True, text=True)
print(result.stdout)
</action>

<action>
import subprocess, os, glob

# Find essay docx
files = glob.glob('/home/user/*.docx')
essay = [f for f in files if 'reminder' not in f][0]
print(essay)

# Convert to PDF
subprocess.run(['soffice', '--convert-to', 'pdf', essay, '--outdir', '/home/user/'], check=True)

# Create zip with docx and pdf
pdf = essay.replace('.docx', '.pdf')
subprocess.run(['zip', '-j', '/home/user/Desktop/essay_submission.zip', essay, pdf], check=True)
print("Done")
</action>

===== Step 2 (predict_num=2, phase=act) =====
I can see the requirements clearly:
1. Export essay as PDF
2. Pack the writer file (docx) along with the pdf into a ZIP named 'essay_submission.zip'

Let me open a terminal and execute the commands:

<action>
import pyautogui, time
pyautogui.hotkey('ctrl', 'alt', 't')
time.sleep(2)
</action>

<action>
import pyautogui, time
# Find essay docx files (excluding reminder.docx)
pyautogui.typewrite('ls /home/user/*.docx', interval=0.05)
pyautogui.press('enter')
time.sleep(1)
</action>

<action>
import pyautogui, time
# List candidate docx files
pyautogui.typewrite('cd /home/user && for f in *.docx; do if [ "$f" != "reminder.docx" ]; then echo "$f"; fi; done', interval=0.05)
pyautogui.press('enter')
time.sleep(1)
</action>

<action>
import pyautogui, subprocess, glob, os, time

files = glob.glob('/home/user/*.docx')
essay = [f for f in files if 'reminder' not in os.path.basename(f).lower()][0]
print("Essay:", essay)

result = subprocess.run(['soffice', '--headless', '--convert-to', 'pdf', essay,
                         '--outdir', '/home/user/'], capture_output=True, text=True)
print(result.stdout); print(result.stderr)

pdf = os.path.splitext(essay)[0] + '.pdf'
print("PDF:", pdf, "exists:", os.path.exists(pdf))

zip_path = '/home/user/Desktop/essay_submission.zip'
result2 = subprocess.run(['zip', '-j', zip_path, essay, pdf], capture_output=True, text=True)
print(result2.stdout); print(result2.stderr)
print("ZIP created:", os.path.exists(zip_path))
</action>

===== Step 3 (predict_num=3, phase=act) =====
I can see the essay file is `Recruitment_..._Europe.docx`. Now I need to convert it to PDF and create the zip file.

<action>
import pyautogui, time
# Click on the terminal
pyautogui.click(310, 177)
time.sleep(0.3)
# Convert the essay to PDF
pyautogui.typewrite('soffice --headless --convert-to pdf /home/user/Recruitment_..._Europe.docx --outdir /home/user/', interval=0.05)
pyautogui.press('enter')
time.sleep(5)
</action>

<action>
import pyautogui, time
# Create the zip with both the docx and pdf files
pyautogui.typewrite('zip -j /home/user/Desktop/essay_submission.zip /home/user/Recruitment_..._Europe.docx /home/user/Recruitment_..._Europe.pdf', interval=0.05)
pyautogui.press('enter')
time.sleep(2)
</action>

<action>
import pyautogui, time
# Verify the zip was created
pyautogui.typewrite('ls -la /home/user/Desktop/essay_submission.zip', interval=0.05)
pyautogui.press('enter')
time.sleep(1)
# Also verify contents
pyautogui.typewrite('unzip -l /home/user/Desktop/essay_submission.zip', interval=0.05)
pyautogui.press('enter')
time.sleep(1)
</action>

===== Step 4 (predict_num=4, phase=act) =====
The task is complete. I can see that:

1. The essay was successfully converted to PDF (`Recruitment_..._Europe.pdf`)
2. The ZIP file `essay_submission.zip` was created at `/home/user/Desktop/essay_submission.zip` (279966 bytes)
3. The ZIP contains both files:
   - `Recruitment_..._Europe.docx` (168712 bytes)
   - `Recruitment_..._Europe.pdf` (121162 bytes)

All requirements from reminder.docx have been met:
- Essay exported as PDF format
- Writer file (docx) and PDF packed into ZIP named 'essay_submission.zip'
- ZIP placed on Desktop

<action>DONE</action>
\end{failbox}

\begin{failbox}{Analyzer Root-Cause Diagnosis --- false-positive \texttt{likely\_success}}
<root_cause>
  <trajectory_summary>
The agent read the reminder.docx which specified two requirements: (1) export the
essay as a PDF, and (2) pack the writer file (docx) along with the PDF into a ZIP
named 'essay_submission.zip'. The agent identified the essay file as
Recruitment_..._Europe.docx in /home/user/.
It then used LibreOffice headless mode to convert the docx to PDF, and used the zip
command to pack both files into /home/user/Desktop/essay_submission.zip. The agent
verified the result using ls and unzip -l, confirming the ZIP contains both the
docx (168712 bytes) and PDF (121162 bytes). The agent then declared the task complete.
  </trajectory_summary>
  <what_worked>
    <item>Turn 2: The agent correctly identified the essay file by listing
    /home/user/*.docx and filtering out reminder.docx, finding
    Recruitment_..._Europe.docx.</item>
    <item>Turn 3: The LibreOffice headless conversion succeeded -- the terminal
    showed the soffice command running and the PDF was produced at
    /home/user/Recruitment_..._Europe.pdf.</item>
    <item>Turn 3: The zip command successfully created
    /home/user/Desktop/essay_submission.zip containing both the docx and pdf
    files, confirmed by the terminal output showing "adding: ... updating: ..."
    and the final file size of 279966 bytes.</item>
    <item>Turn 3: The unzip -l verification confirmed the ZIP contains exactly 2
    files: the .docx and .pdf, matching both requirements from reminder.docx.</item>
  </what_worked>
  <issues>
  </issues>
  <outcome_assessment value="likely_success">The terminal output in Turn 3 clearly
  shows essay_submission.zip was created at /home/user/Desktop/ containing both the
  docx and the PDF, satisfying all requirements stated in reminder.docx.</outcome_assessment>
</root_cause>
\end{failbox}

\subsection{MMG2Skill-Game: Craft Stick}
\label{app:case_bamboo}

\noindent\textbf{Guide-supplied procedural knowledge.} This case provides the trajectory evidence behind the bamboo-to-sticks example in \S\ref{sec:exp:main}. The vanilla agent searches for wooden planks and exhausts the budget, while MMG2Skill uses the guide-derived bamboo recipe and completes the GUI crafting sequence. This isolates a guide-supplied recipe rather than a larger search budget.

\begin{instructionbox}
\textbf{Task instruction.} Craft a stick.
\end{instructionbox}

\begin{failbox}{Vanilla Agent Trajectory (\texttt{craft\_item:stick\_zero}, \texttt{claude-sonnet-4.6})}
===== Step 1 (predict_num=1, phase=act) =====
I can see the inventory/crafting GUI is open. I need to craft a stick. To craft sticks, I need to place 2 planks vertically in the crafting grid. Looking at the inventory, I can see there are some items. I need to find wooden planks. I can see what appears to be some items in the inventory - there's a stack of 29 items that looks like it could be planks or similar.

The crafting grid is the 2x2 grid in the upper right area. I need to place planks in the top slot and middle slot of one column to make sticks.

Let me first move to the item in the inventory that appears to be planks (the stack of 29). The cursor appears to be near the center of the screen. I need to move to that inventory item first to identify it, then place it in the crafting grid.

The stack of 29 items is in the inventory row - it appears to be in the second row of the inventory, second slot from the left. Let me move there to pick it up.

<action>move(-25, 35) and click(left)</action>

===== Step 2 (predict_num=2, phase=act) =====
The inventory appears empty - no planks visible. I need to close this and get some wood first. Let me close the inventory and mine a log.

<action>press(e)</action>

(*@{\normalfont\itshape\color{gray!130}[\ldots\ agent keeps searching for wooden planks until the 60-step budget is exhausted.]}@*)
\end{failbox}

\begin{successbox}{MMG2Skill Agent Trajectory (\texttt{craft\_item:stick\_zero}, \texttt{claude-sonnet-4.6})}
===== Step 1 (predict_num=1, phase=act) =====
I see bamboo (32) in inventory. I'll click on it, then place in crafting grid vertically.

<action>move(-55, 60) and click(left)</action>

===== Step 2 (predict_num=2, phase=act) =====
I can see the bamboo stack (32) is in the inventory. The cursor is near the bottom-left area. I need to click on the bamboo stack to pick it up.

<action>move(27, -27) and click(left)</action>

===== Step 3 (predict_num=3, phase=act) =====
The cursor is hovering over the bamboo stack (32) but I haven't picked it up yet. I need to click on it to pick it up.

<action>click(left)</action>

(*@{\normalfont\itshape\color{gray!130}[\ldots\ agent positions bamboo into the 2$\times$2 crafting grid through a sequence of GUI clicks.]}@*)

===== Step 29 (predict_num=29, phase=act) =====
The cursor is now right next to the output slot (just to the left of it). The stick is visible in the output slot. I need to move slightly right to be over the output slot and shift-click it.

<action>move(15, 0) and press(left.shift) and click(left)</action>

===== Step 30 (predict_num=30, phase=act) =====
The crafting was successful! I can see "New Recipes Unlocked!" notification, the crafting grid is now empty, and there's a stick visible in the bottom-right of the inventory. The stick has been crafted and collected into the inventory. The task is complete.

<action>DONE</action>
\end{successbox}

\subsection{MMG2Skill-Game: Craft Wheat}
\label{app:case_wheat}

\noindent\textbf{Guide-supplied recipe correction.} This case is a complementary instance of the bamboo-to-sticks pattern (\S\ref{sec:exp:main}). The vanilla agent maps wheat only to harvesting and emits \texttt{FAIL} despite having hay bales in inventory, while MMG2Skill uses the guide-derived hay-bale-to-wheat reverse recipe and collects the wheat output. The case shows that guide-derived skills recover non-obvious recipes as well as long action sequences.

\begin{instructionbox}
\textbf{Task instruction.} Craft wheat.
\end{instructionbox}

\begin{failbox}{Vanilla Agent Trajectory (\texttt{craft\_item:wheat\_zero}, \texttt{claude-sonnet-4.6})}
===== Step 1 (predict_num=1, phase=act) =====
I can see the crafting table is open. The inventory shows what appears to be wheat seeds (the yellow item with 39 count) in the bottom row. However, wheat cannot be crafted - it's obtained by harvesting wheat crops. Wait, let me reconsider. The task says "craft wheat." In Minecraft, wheat cannot be crafted from other items in the standard recipe. However, wheat CAN be crafted - actually no, wheat is obtained by harvesting. But wait - looking more carefully at the inventory, I see a yellow item that looks like it could be wheat seeds or wheat. The inventory appears mostly empty except for that one stack. Actually, in Minecraft there's no crafting recipe for wheat itself. Wheat is obtained by harvesting wheat crops. This task might be impossible to complete through crafting.

But let me reconsider - maybe the task means something else, or perhaps I'm missing something in the inventory. The inventory shows only one item visible (yellow, 39 count in the bottom row). Since wheat cannot be crafted in Minecraft (there's no recipe that produces wheat), this task cannot be completed.

<action>FAIL</action>
\end{failbox}

\begin{successbox}{MMG2Skill Agent Trajectory (\texttt{craft\_item:wheat\_zero}, \texttt{claude-sonnet-4.6})}
===== Step 1 (predict_num=1, phase=act) =====
I can see the inventory GUI is open with a 2x2 crafting grid. I can see there's an item in the bottom row of the inventory - it appears to be a hay bale (the yellowish item with "39" count). I need to pick it up and place it in the crafting grid.

The cursor is currently near the center of the inventory. I need to move to the hay bale item in the bottom inventory row (the one with the golden/yellow item showing count 39).

<action>move(30, 90)</action>

===== Step 2 (predict_num=2, phase=act) =====
The cursor moved too far down and is now below the GUI. I can see the hay bale item in the bottom row of the inventory (the yellowish item with count 39). I need to move the cursor up to that hay bale slot.

<action>move(-20, -50)</action>

(*@{\normalfont\itshape\color{gray!130}[\ldots\ agent positions the hay bale into the crafting grid and harvests the output through a sequence of GUI clicks.]}@*)

===== Step 39 (predict_num=39, phase=act) =====
The tooltip shows "Wheat" - the cursor is on the output slot! Now shift-click to collect the wheat.

<action>press(left.shift) and click(left)</action>

===== Step 40 (predict_num=40, phase=act) =====
The wheat has been crafted and collected! I can see:
- "New Recipes Unlocked! Check your recipe book" notification appeared
- The crafting grid is now empty (hay bale consumed)
- The output slot is empty
- In the bottom row of the inventory, I can see a wheat stack with "9" in the last slot
- Also 35 hay bales remain in inventory

The wheat (9 pieces) has been successfully crafted and moved to my inventory. The task is complete!

<action>DONE</action>
\end{successbox}

\section{Prompt Templates}
\label{app:prompts}

We reproduce the system-level prompts used by the three MMG2Skill operators in Algorithm~\ref{alg:mmg2skill_full}. Runtime placeholders such as \texttt{\{domain\_guidance\}} and \texttt{\{domain\_reviser\_guidance\}} are filled with benchmark-specific conventions.

\subsection{Skill construction prompt}
\label{app:prompts:construct}

\textsc{ConstructSkills} receives the task instruction and normalized guide content, then emits one or more self-contained skills in Markdown.

\begin{promptbox}{Skill Construction Prompt}
You are an expert at analyzing tutorials and extracting reusable skills.
The tutorial may be provided as raw HTML, markdown, or plain text --
extract the meaningful content regardless of format. Ignore navigation,
ads, scripts, sidebars, and other non-tutorial noise.
Given a tutorial and a specific task instruction, extract one or more
skills -- each a self-contained procedure for a sub-task. Together, the
skills should cover all steps needed to accomplish the task.
{domain_guidance}
Output each skill as a markdown section with this format:
# skill-name-in-kebab-case
> Brief description of what this skill does (1-2 sentences)
## Steps
1. Right-click on the desktop background
   ![Right-click context menu](screenshot-3.png)
2. In the settings panel, toggle Bluetooth on
   ![Settings panel showing Bluetooth toggle](screenshot-5.png)
## Expected Result
What the result should be after completing this skill.
---
Rules:
- Use # (h1) for the skill name, in kebab-case
- Use > blockquote for the description; separate skills with ---
- Use ![brief description](filename) after key steps to reference
  tutorial images. ONLY use filenames from [Image: ...] labels.
- Be specific about actions and expected outcomes
- Each step should be a single atomic action
\end{promptbox}

\subsection{Analyzer prompt}
\label{app:prompts:analyzer}

The analyzer reads trajectory chunks and emits structured XML evidence for downstream stopping and refinement decisions. We show the final-round prompt, which produces the \texttt{<root\_cause>} block used by Algorithm~\ref{alg:mmg2skill_full}.

\begin{promptbox}{Analyzer Prompt (Phase 1a --- Intermediate Round)}
You are a trajectory analyst reading the execution trace of an AI agent.

{domain_reviser_guidance}

## How the trace is fed to you
The trace is delivered in fixed-size windows of predict-turns (one turn
= one agent LLM call). In each round you receive:
- the task instruction,
- (if any) a rolling summary of earlier turns,
- on the FIRST window only, the initial observation the agent
  perceived before it made any action — this is NOT a turn; do not
  index it as Turn 0 in `<where>`, only use it to ground the agent's
  opening decision,
- the detailed content of the current window: for every turn, the
  agent's own response (possibly truncated), the action(s) it
  emitted, and the saved observation for the turn's last action.

The saved observation is whatever the benchmark kit returns — it may
be an image, a block of text, or both, depending on what the agent
actually perceives in that domain.

You see exactly what the agent itself saw or wrote — no controller
errors, return codes, or env-side `done` flags. If an action failed
silently, the only evidence is the post-action observation, just like
the next agent's evidence will be.

## Your job this round
This is NOT the final round. Update the rolling summary so the next
round can continue the analysis without re-reading earlier turns.

Focus on:
- what the agent attempted and whether it succeeded,
- state transitions that matter (GUI opened / closed, inventory changed,
  new error emerged, etc.),
- recurring patterns (same failure loop, oscillating corrections, etc.),
- local sub-steps that clearly *worked* — the next round will need to
  know what to preserve, not only what went wrong.

Note: if the agent emits any terminal "cannot finish / task infeasible"
signal, treat it as a hypothesis from the agent, not a verdict. Record
the agent's justification verbatim so the next round can check it
against the observations — some tasks are genuinely infeasible and
bailing out is the correct answer. The exact syntax for such signals
is benchmark-specific; rely on the agent's wording and any domain hints
above to recognise them.

## Output format
Emit EXACTLY one block, nothing else:

<summary>
Plain prose describing the trajectory so far. Reference turns by index.
Keep it under {rolling_summary_char_limit} characters.
</summary>

Rules:
- Base the summary only on what is visible in the trace. Do NOT invent
  external context or information not shown.
- No markdown, no JSON, no preamble before the tag. Just `<summary>...</summary>`.
\end{promptbox}

\begin{promptbox}{Analyzer Prompt (Phase 1b --- Final Round)}
You are a trajectory analyst reading the final window of an AI agent's
execution trace. Your job is to produce a root-cause analysis as XML.

{domain_reviser_guidance}

## How the trace is fed to you
The task instruction, (optionally) a rolling summary of earlier turns,
and the detailed content of the final window: for every turn, the
agent's own response (possibly truncated), the action(s) it emitted,
and the saved observation for the turn's last action. If this is also
the first window (single-chunk run), you additionally receive the
initial observation before Turn 1 — that block is NOT a turn; do not
index it as Turn 0 in `<where>`, only use it to ground the agent's
opening decision.

The saved observation is whatever the benchmark kit returns — it may
be an image, a block of text, or both, depending on what the agent
actually perceives in that domain.

You see exactly what the agent itself saw or wrote — no controller
errors, return codes, or env-side `done` flags. If an action failed
silently, the only evidence is the post-action observation, just like
the next agent's evidence will be.

## Judging the outcome (no grader available)
You do NOT have access to any rule-based grader. If the agent emitted
a terminal "done / task completed" signal, that is a self-claim — do
not assume the task actually succeeded just because the agent said so.
Conversely, a terminal "cannot finish / task infeasible" signal is a
hypothesis, not a verdict: some tasks are genuinely infeasible
(missing prerequisite, resource unavailable, contradictory instruction,
required tool not installed, etc.) and bailing out with well-grounded
reasoning is the *correct* output. The exact syntax of these terminal
signals is benchmark-specific; recognise them from the agent's wording
and any domain hints above, not from any assumed token.

You must therefore form your own judgment from the visible evidence.
Walk through the trace sub-step by sub-step: which intermediate goals
reached verifiable state, which did not, and whether the final
observation is consistent with the task having been completed.

## What to emit
A `<root_cause>` XML block describing:
1. a concise `<trajectory_summary>` of the whole run (what the agent
   attempted, the path it took, the final state it left behind);
2. `<what_worked>` — specific sub-steps that were executed correctly,
   listed as separate `<item>` entries. These must stay stable across
   refinements, so be concrete: quote the action / observation that
   evidences the success. Populate this even when the overall outcome
   is `likely_failure` — local correctness matters for preservation.
3. `<issues>` — concrete problems you observed, with `<where>`,
   `<evidence>`, and `<cause>`. May be empty if nothing is broken.
4. a final `<outcome_assessment>` judging the run as a whole. Put this
   LAST so your verdict follows from the evidence you just listed,
   not the other way around.

## Output format
Emit EXACTLY one block, nothing else:

<root_cause>
  <trajectory_summary>
  Plain prose summary of the trajectory. Reference turns by index where
  helpful. Keep under {rolling_summary_char_limit} characters.
  </trajectory_summary>
  <what_worked>
    <item>Concrete sub-step that was done correctly, grounded in evidence (turn index + quote / observation).</item>
    <!-- 0 or more item blocks. -->
  </what_worked>
  <issues>
    <issue>
      <where>turn 3 / turns 5-7 / etc.</where>
      <evidence>What you saw in the trace (quote or paraphrase the response, action, or observation). Be specific.</evidence>
      <cause>Why that behavior happened, inferred from the evidence.</cause>
    </issue>
    <!-- 0 or more issue blocks. If the run looks clean, it is OK to
         emit zero issues. -->
  </issues>
  <outcome_assessment value="likely_success">One sentence justifying the chosen value, grounded in the evidence above.</outcome_assessment>
</root_cause>

Rules:
- `outcome_assessment value` MUST be exactly one of
  `likely_success`, `uncertain`, `likely_failure`. Use `uncertain`
  when the evidence is mixed or you cannot confidently tell whether
  the task completed — do NOT force a binary verdict.
- Base every claim on concrete evidence from the trace (turn index,
  response quote, action, or observation).
- Use plain text inside tags. No markdown, no JSON, no CDATA sections.
- Do not emit anything before `<root_cause>` or after `</root_cause>`.
\end{promptbox}

\subsection{Refiner prompt}
\label{app:prompts:refiner}

The refiner receives the current skills, guide, and analyzer feedback, then rewrites the skill file in-place while preserving validated behavior.

\begin{promptbox}{Refiner Prompt}
You are a skill-refinement expert. You will be given:
- the task instruction,
- the current SOP skills the agent used,
- the full tutorial the skills were extracted from (text + images),
- a `<root_cause>` XML analysis of the agent's most recent attempt,
- (if this is not attempt 2) the `<root_cause>` XML of every prior
  attempt in chronological order, so you can see how the skills have
  already been edited and avoid undoing fixes from earlier rounds.

{domain_reviser_guidance}

## Your job
Rewrite the skills so the next attempt avoids the issues flagged in
the most recent `<root_cause>` while preserving the sub-steps listed
in `<what_worked>`. Use the tutorial as the source of truth — if the
skill drifts from what the tutorial actually says, realign it.
Preserve skills that the analysis does not implicate.

## Editing style (applies regardless of outcome)
Refinement is **in-place editing of existing skills**. Do not grow
the skill set, and avoid growing any individual skill unless strictly
necessary — bloated skills lose the agent's attention and are the
main way previous fixes get forgotten.

### Handling `<issues>`
- Prefer to **rewrite** the step that produced the issue so the bad
  behaviour is no longer the natural reading. Do NOT address an issue
  by appending a cautionary sentence after the step; fix the step
  itself.
- You may add a new step ONLY when the issue is a genuinely missing
  action (e.g., the agent skipped a necessary wait, or skipped a
  verification that the tutorial shows). Do not add new steps for
  "just in case" hardening.
- Do not split one issue's fix across multiple skills, and do not
  create a new skill just to house a fix.

### Handling `<what_worked>`
- Treat each item as empirical evidence of a concrete method
  (shortcut, command flag, tool choice) that actually worked in this
  run.
- If a relevant skill is **silent** about that sub-step, promote the
  observed method into the most relevant existing skill so future
  attempts bias toward it.
- If a relevant skill already prescribes a **different** method, you
  may overwrite it with the observed method ONLY when the tutorial is
  itself silent or ambiguous about which to use. Tutorial wins on
  direct conflict — never replace a tutorial-backed method with the
  observed one.
- Do not create a new skill to carry `<what_worked>` content, and do
  not duplicate the same method across multiple skills.

## Edit intensity is gated by <outcome_assessment>
The most recent `<root_cause>` carries an `<outcome_assessment
value="...">` on its last line. Honor it:

- `value="likely_success"` — the analyzer judged the run effectively
  correct. Minimal mode:
  * if `<issues>` is empty you are in **reinforce mode**: absorb each
    `<what_worked>` item into the most relevant silent skill per the
    `<what_worked>` rule above, and make no other edits;
  * otherwise only tighten wording in skills explicitly pointed to by
    `<issues>`, and for `<what_worked>` absorb a method ONLY when the
    relevant skill is silent about that sub-step — do NOT overwrite a
    skill's existing method even if the tutorial is ambiguous.
- `value="uncertain"` — normal editing: apply the Editing style rules
  above for both `<issues>` and `<what_worked>` without further
  restriction.
- `value="likely_failure"` — larger in-place edits allowed: you may
  reshape steps and add a missing verification / wait when an issue
  calls for it. Actively absorb `<what_worked>` so the local wins do
  not regress. Still no new skills, still in-place.

## Using history
When prior attempts' `<root_cause>` XMLs are provided:
- If the current `<issues>` would undo a fix an earlier attempt
  introduced, prefer a different repair over reverting prior work.
- If a `<what_worked>` method was already absorbed into a skill in an
  earlier round (visible in the current skills), do not re-absorb it.

## Output format
Return the refined skills in the same markdown shape that the upstream
skill extractor produces:

# skill-name-in-kebab-case
> Brief description of what this skill does

## Steps
1. First step
2. Second step
   ![context image](filename.png)

## Expected Result
What the result should be.

---

# another-skill
> ...

Rules:
- Keep skill names in kebab-case; `> description` on its own line.
- Separate skills with `---` on a line by itself.
- You may reference tutorial images inline with `![alt](filename.png)`,
  where filename matches one of the image filenames shown in the user
  message. Images referenced in the original skills are also valid.
- Do NOT fabricate totally new skills unless the tutorial demonstrates
  a path that is genuinely missing. Prefer editing existing skills.
- Every skill must be present exactly once in the output (no dropping
  untouched skills — pass them through unchanged).
- No JSON, no outer code fences, no preamble before the first `#`.
\end{promptbox}

\subsection{Agent prompts}
\label{app:prompts:agents}

We additionally reproduce the system prompts used by the agent at rollout time, for each of the three configurations compared in Section~\ref{sec:exp:setup}: \texttt{vanilla} (no external skills), \emph{Raw Guide} (raw guide injected), and the MMG2Skill agent (consumes the constructed/refined \texttt{SKILL.md}).

\subsubsection{Vanilla agent prompt}
\label{app:prompts:vanilla}
\begin{promptbox}{Vanilla Agent Prompt}
{domain_system_prompt}

Think through your observations and reasoning before producing your action.
\end{promptbox}

\subsubsection{Raw Guide agent prompt}
\label{app:prompts:vanilla_tutorial}
\begin{promptbox}{Vanilla + Tutorial Agent Prompt}
{domain_system_prompt}

You are given the original tutorial document (text/HTML plus images) as
reference material. This is RAW source material, not a structured skill —
it may contain navigation, ads, irrelevant sections, or steps that don't
match the actual environment (different versions, layouts, prerequisites).

On each turn:
1. OBSERVE the current state carefully — what do you actually see in the current observation?
2. EXTRACT relevant procedural knowledge from the tutorial yourself —
   identify the steps that apply to your current state and ignore the rest.
3. ASSESS whether the tutorial's procedure matches what you actually see;
   adapt or deviate when it doesn't.
4. DECIDE: if the overall task is fully complete, signal completion using
   the domain-specified format. Otherwise generate the next action based
   on the current state — not by blindly replaying the tutorial.
5. HANDLE ERRORS naturally: diagnose from the actual observation rather
   than re-reading the tutorial.

Think through your observations and reasoning before producing your action.
\end{promptbox}

\subsubsection{MMG2Skill agent prompt}
\label{app:prompts:mmg2skill_agent}
\begin{promptbox}{MMG2Skill Agent Prompt}
{domain_system_prompt}

You have reference skills extracted from tutorials. These skills are for REFERENCE
ONLY — they may not match the actual environment (different OS version, different
UI layout, missing prerequisites, etc.). You must judge how much to rely on them
based on what you actually observe.

Your responsibilities on each turn:
1. OBSERVE the current state carefully — what do you actually see in the current observation?
2. ASSESS whether the reference skills apply: does the current state match what
   the SOP assumes? If not, adapt or ignore the SOP and act on what you see.
3. DECIDE: if the overall task is fully complete, signal completion using the
   domain-specified format. Otherwise, generate the next action based on the
   current state — not by blindly following SOP steps.
4. HANDLE ERRORS naturally: if something unexpected happened, diagnose from the
   actual observation and try a different approach rather than repeating the SOP.

Think through your observations and reasoning before producing your action.
\end{promptbox}

\end{document}